\documentclass[10pt]{article} 
\usepackage[preprint]{tmlr}
\usepackage{makecell}
\usepackage{hyperref}
\usepackage{url}
\usepackage{times}
\usepackage{latexsym}
\usepackage{graphicx}
\usepackage{inconsolata}
\usepackage{graphicx}
\usepackage{amsmath}
\usepackage{booktabs}
\usepackage{arydshln}
\usepackage{multirow}
\usepackage{wrapfig, lipsum, booktabs}
\usepackage{amsfonts}
\usepackage{algorithm}
\usepackage{enumerate}
\usepackage{enumitem}
\usepackage[T1]{fontenc}
\usepackage{fdsymbol}

\usepackage[utf8]{inputenc}

\usepackage{microtype}

\usepackage{inconsolata}
\title{Igniting Language Intelligence: The Hitchhiker's Guide From Chain-of-Thought Reasoning to Language Agents
}

\author{\name Zhuosheng Zhang$^{\clubsuit,}$\thanks{Equal contribution. We thank Diyi Yang for providing valuable feedback on the draft.} , Yao Yao$^{\clubsuit,*}$, Aston Zhang$^\varheartsuit$, Xiangru Tang$^\spadesuit$, Xinbei Ma$^\clubsuit$, Zhiwei He$^\clubsuit$, Yiming Wang$^\clubsuit$, Mark Gerstein$^\spadesuit$, Rui Wang$^\clubsuit$, Gongshen Liu$^\clubsuit$, Hai Zhao$^\clubsuit$\\
\email \{zhangzs,yaoyao27,sjtumaxb,zwhe.cs,wangrui12,lgshen\}@sjtu.edu.cn,
az@astonzhang.com,
\{xiangru.tang,mark.gerstein\}@yale.edu,
alsaceym@gmail.com,
zhaohai@cs.sjtu.edu.cn
\\
 \addr $^\clubsuit$Shanghai Jiao Tong University, $^\varheartsuit$Amazon Web Services, $^\spadesuit$Yale University
}

\def\month{MM}  
\def\year{YYYY} 
\def\openreview{\url{https://openreview.net/forum?id=XXXX}} 

\begin{document}
\maketitle

\begin{figure*}[htb]
    \centering
    \vspace{-3mm}
    \includegraphics[width=1.0\textwidth]{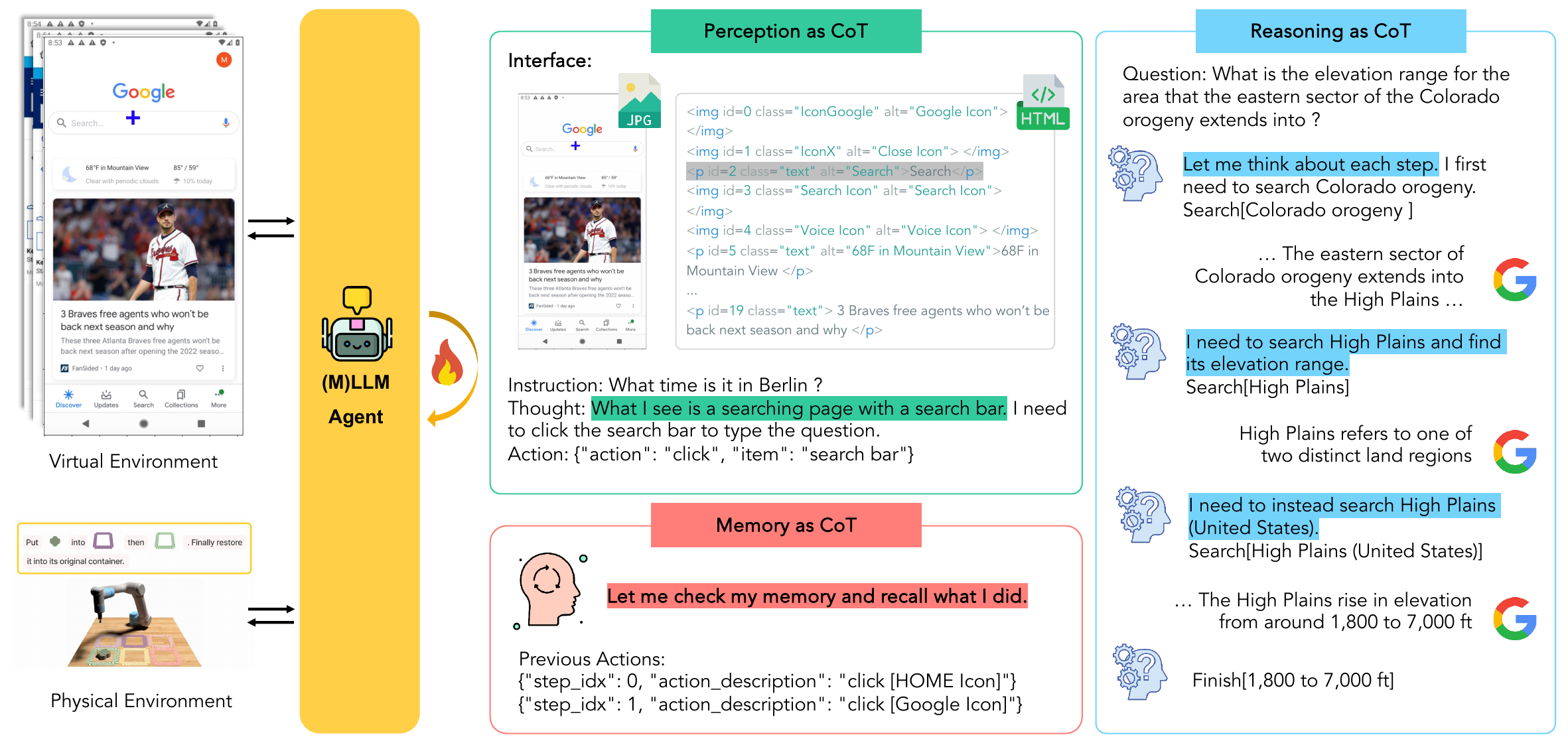}
    \vspace{-6mm}
    \caption{An overview of language agent framework empowered with the chain-of-thought (CoT) mechanism in perception, memory, and reasoning.}
    \label{fig:cot-role}
    \vspace{-1mm}
\end{figure*}

\begin{abstract}
Large language models (LLMs) have dramatically enhanced the field of language intelligence, as demonstrably evidenced by their formidable empirical performance across a spectrum of complex reasoning tasks. Additionally, theoretical proofs have illuminated their emergent reasoning capabilities, providing a compelling showcase of their advanced cognitive abilities in linguistic contexts.
Critical to their remarkable efficacy in handling complex reasoning tasks, LLMs leverage the intriguing chain-of-thought (CoT) reasoning techniques, obliging them to formulate intermediate steps en route to deriving an answer. The CoT reasoning approach has not only exhibited proficiency in amplifying reasoning performance but also in enhancing interpretability, controllability, and flexibility. In light of these merits, recent research endeavors have extended CoT reasoning methodologies to nurture the development of autonomous language agents, which adeptly adhere to language instructions and execute actions within varied environments.
This survey paper orchestrates a thorough discourse, penetrating vital research dimensions, encompassing: (i) the foundational mechanics of CoT techniques, with a focus on elucidating the circumstances and justification behind its efficacy; (ii) the paradigm shift in CoT; and (iii) the burgeoning of language agents fortified by CoT approaches. Prospective research avenues envelop explorations into generalization, efficiency, customization, scaling, and safety. We hope to offer readers a comprehensive understanding of prevalent research areas such as CoT reasoning and language agents and illuminate the interconnections weaving through these areas. 
This paper caters to a wide audience, including beginners seeking comprehensive knowledge of CoT reasoning and language agents, as well as experienced researchers interested in foundational mechanics and engaging in cutting-edge discussions on these topics.
A repository for the related papers is available at 
\url{https://github.com/Zoeyyao27/CoT-Igniting-Agent}.

\end{abstract}

\section{Introduction}

Language intelligence pertains to the aptitude for comprehending and reasoning through concepts articulated in natural languages
\citep{STERNBERG1982155,ryan2001wechsler,ramsden2011verbal,luwel2013role}.
Spurred by advancements in scale, large language models (LLMs) have achieved remarkable progress in pursuing human-level language intelligence, compellingly evidenced by the strong empirical benchmarking performance in complex reasoning tasks \citep{wei2022emergent}, as well as theoretical proofs for the emergent reasoning abilities \citep{prystawski2023think,wang2023reasoning,bi2023program}.

Reasoning, a pivotal research topic within the realm of language intelligence, is characterized as a multi-step process wherein inferences are drawn from discrete pieces of evidence, culminating in the formation of more abstract concepts that are instrumental in facilitating high-level predictions \citep{reasoning,yu2023nature,huang-chang-2023-towards}. Recent research revealed that remarkable enhancements in performance could be attained by prompting LLMs to engage in a step-by-step reasoning process, as opposed to generating answers in a direct manner \citep{scratchpad,wei2023chainofthought}. The way to prompt LLMs to generate a series of intermediate reasoning steps for solving a problem is called chain-of-thought (CoT) prompting \citep{wei2023chainofthought}. 

Optimization of CoT prompting techniques has garnered escalating interest, catalyzing notable paradigm shifts within the CoT framework. These shifts encompass three key aspects: 
(i) \textit{prompting pattern}: from manual design of in-context learning demonstrations to automatic prompt construction \citep{zhang2023automatic,zhou2022large,yang2023large}; 
(ii) \textit{reasoning format}: from unstructured natural language formats to structured ones \citep{chen2022program,ziqi2023tab,yao2023tree,DBLP:journals/corr/abs-2305-16582}; 
(iii) \textit{application scenario}: from singular language settings to multilingual environments \citep{shi2022language}, from a language modality to embracing multimodal approaches \citep{zhang2023multimodal}, and from complex reasoning tasks to general-purpose tasks \citep{wang2022rationaleaugmented, li2022self, wang-etal-2023-element, he2023exploring}.

\begin{figure*}[t]
    \centering
\includegraphics[width=1\textwidth]{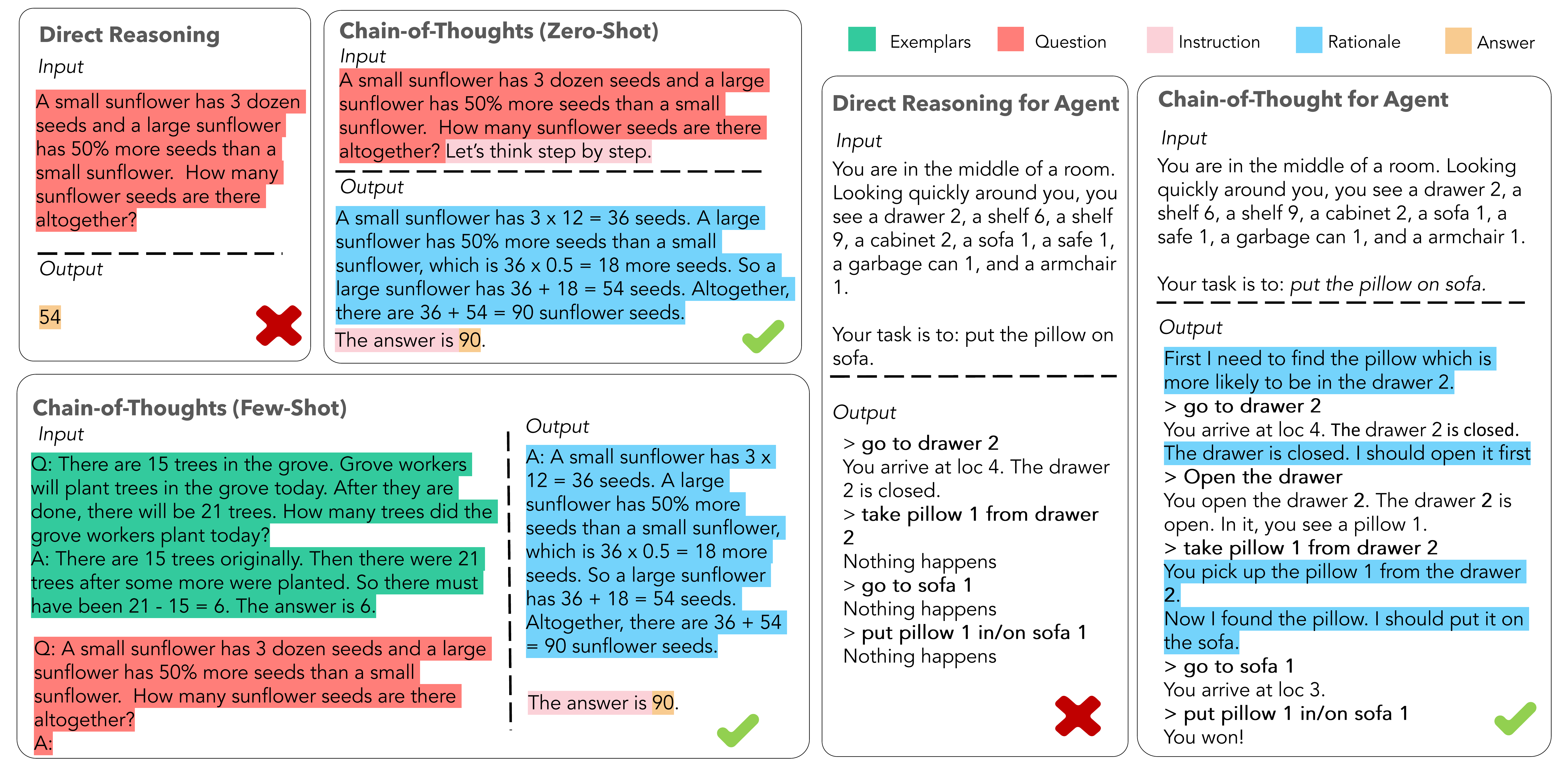}
    \caption{Comparison between CoT reasoning and direct reasoning. CoT refers to a series of intermediate reasoning steps that are generated to solve a problem or arrive at an answer  ~\citep{wei2023chainofthought}. This approach is often more effective than direct reasoning, which attempts to tackle the entire problem all at once. }
\label{fig:cot_vs_direct}
\end{figure*}

CoT reasoning is a representative emergent ability of LLMs \citep{wei2022emergent}. It provides a proficient strategy for deconstructing intricate issues into smaller, manageable sub-problems, systematically enabling solutions through a step-by-step approach (Figure \ref{fig:cot_vs_direct}). 
Leveraging the reasoning capabilities developed during pre-training \citep{xie2022explanation,wang-etal-2023-towards}, CoT prompting adeptly identifies atomic knowledge components essential for reasoning processes and seamlessly integrates their relationships, thereby constructing intermediate, coherent reasoning steps \citep{prystawski2023think,wang2023reasoning}. 
In addressing these sub-problems, the reasoning process can be further enhanced by employing knowledge retrieval and verification tools \citep{gou2023critic, qin2023toolllm}. By expanding CoT into a comprehensive framework for perception, memory, and reasoning, language agents, powered by LLMs, have been formulated to adeptly adhere to language instructions and execute actions in either real-world or simulated environments
\citep{rawles2023android,zhang2023you} (Figure \ref{fig:cot-role}). These language agents come in two flavors: (i) autonomous agents \citep{act-1,autogpt,hong2023metagpt,babyagi} and (ii) communicative agents \citep{park2023generative,wang2023voyager,zhu2023ghost,hong2023metagpt}.

In this paper, we navigate through research topics that encompass: (i) unraveling the underlying mechanisms of CoT techniques, with a particular focus on discerning when and why CoT is effective; (ii) identifying and analyzing the paradigm shift occurring within CoT; and (iii) investigating the advent of language agents enabled by CoT techniques. The rest of this paper is structured for a coherent and sequential exploration. Initially, we immerse ourselves in the fundamental aspects of CoT reasoning, which encompasses its defining features and the merits arising from employing CoT techniques. Subsequently, we delve deeper into the inherent mechanisms of CoT, striving to elucidate the specific conditions and reasons that determine its functionality. In the ensuing section, we classify paradigm shifts, directing our attention towards various prompting techniques, reasoning formats, and application scenarios. Following that, we explore the emerging landscape of language agents, spotlighting those facilitated by CoT techniques. To conclude, we engage in a discussion about the challenges encountered and future opportunities looming on the horizon. 

Various related papers have selectively concentrated on distinct facets of LLMs \citep{zhao2023survey}, CoT reasoning \citep{lu2022survey,qiao2022reasoning,chu2023survey,yu2023towards}, and autonomous agents \citep{wang2023survey,xi2023rise}, each providing overviews and taxonomies tailored to their respective domains. In contrast, our paper transcends a mere summarization and aspires to furnish a thorough exploration of the fundamental mechanisms that underscore CoT reasoning, alongside a deep dive into the paradigm shifts enveloping this domain. Moreover, our paper chronicles the trajectory from CoT reasoning in LLMs to the most recent advancements in autonomous language agents, with a dedicated aim to illuminate the intricate interconnections weaving through these crucial areas of study. This paper caters to a wide audience, including beginners seeking comprehensive knowledge of CoT reasoning and language agents, as well as experienced researchers interested in foundational mechanics and cutting-edge discussions on these topics.

\vspace{-2.5mm}
\paragraph{Key Takeaways} To the best of our knowledge, this constitutes the inaugural work to systematically explore the foundational mechanics of CoT techniques, the paradigm shift in CoT, and the complex interplay between CoT and agents. The key takeaways are:

\begin{itemize}[leftmargin=*,topsep=2pt,itemsep=1pt,parsep=0pt]
  \item CoT demonstrates efficacy under two overarching conditions: first, when an LLM with preferably at least 20 billion parameters is employed, and second, when the parametric knowledge within the LLM encompasses knowledge pieces that are (i) pertinent to the task at hand and (ii) maintain strong mutual interconnections (Section \ref{sec:when}).
    \item CoT functions by assisting in the identification of atomic knowledge pieces pivotal for reasoning and seamlessly connecting these components via the formation of intermediate reasoning steps (Section \ref{sec:why}).
    \item CoT techniques have experienced substantial paradigm shifts, embracing alterations in prompting patterns, reasoning formats, and application scenarios (Section \ref{sec:paradigm}).
    \item CoT has acted as a catalyst in the evolution of LLM-empowered agents capable of understanding language instructions and executing actions in both real-world and simulated environments, specifically augmenting agent capabilities in    perception, memory, and reasoning (Section \ref{sec:agents}).
    \item Despite the swift advancement of LLMs, CoT reasoning, and language agents, numerous challenges persist, such as generalization to unseen domains, efficiency amidst redundant interactions, customization of language agents, scaling up of language agents, and ensuring the safety of language agents (Section \ref{sec:challenges}).
\end{itemize}

\section{Preliminaries of CoT}
In this section, we immerse ourselves in the fundamental elements of CoT reasoning. Firstly, we carve out a distinct contrast between CoT reasoning and the traditional approach of direct reasoning. Subsequently, we proffer definitions for the key components within CoT. Finally, we delineate the advantages of adopting CoT.


\subsection{Definition}
The concept of \emph{chain-of-thought} refers to a series of intermediate reasoning steps that are generated to solve a problem or arrive at an answer  ~\citep{wei2023chainofthought}, in the form of <input$\rightarrow$reasoning chain (rationale)$\rightarrow$output> mappings. This approach is often more effective than traditional direct reasoning, which attempts to tackle the entire problem all at once. For example, standard classification, multiple choice, and question answering problems often leverage direct reasoning in the form of <input$\rightarrow$output> mappings.

To elucidate CoT, we establish standard definitions for its key components as illustrated in Figure \ref{fig:cot_vs_direct}.
Formally, assuming that the reasoning dataset distribution is $\mathcal{D}$, we denote $s = (x,y) \sim \mathcal{D}$ as a sampling on $\mathcal{D}$, where $x$ and $y$ denote the question (input) and the answer (output), respectively, and they are both in the form of text sequences. We use $|x|$ to denote the length of the sequence $x$, and $p_{\theta}$ to denote the pre-trained language model parameterized with $\theta$.
Details of definitions are as follows:

\paragraph{Instruction.} 
Instructions are usually short sentences used to prompt an LLM to generate answers in the desired format. They guide the LLM to think step by step in the reasoning process.
We notate the instruction as $\mathsf{p}$, which is set to different text sequences depending on the task requirements.

\paragraph{Rationale.} 
We uniformly refer to the intermediate processes of CoT reasoning as ``rationales''. Rationales can encompass solutions, intermediate reasoning steps, or any relevant external knowledge pertaining to a question. 
We define rationale as $r$. If $r$ is generated by the LLM, instruction $\mathsf{p}$ can be used to obtain $r \sim p_{\theta}(x, \mathsf{p})$. If $r$ is written by a human, instruction $\mathsf{p}$ can be exempted, and $r = f(x)$, where $f(\cdot)$ indicates the handwriting operation.

\paragraph{Exemplars.} 
Exemplars are typically presented as desired input-output pairs in few-shot prompting approaches, each of which contains questions, a rationale, and an answer. Exemplars serve as in-context demonstrations of input-output relationships before generating predictions for test-time examples. Exemplars are usually concatenated before input questions. 
Specifically, assuming the exemplar size of $n$, exemplars $E$ can be formulated as:
\begin{equation}
E = [(x_1, r_1, y_1) \circ \cdots \circ (x_{n-1}, r_{n-1}, y_{n-1}) \circ (x_n, r_n, y_n)],
\end{equation}
where $\circ$ represents concatenation, $(x_i, y_i) \sim \mathcal{D}$ and $r_i = f(x_i)$.

\paragraph{Zero-Shot-CoT.} 
Zero-Shot-CoT does not require users to provide exemplars. Instead, it typically relies on instructions to facilitate the LLM in conducting step-by-step reasoning, thereby generating answers.
For example, \citet{zero-shot} first elicited the LLM to generate the rationale $r$ using the instruction $\mathsf{p_1}$ such as ``\textit{Let's think step by step}'', and then use the instruction $\mathsf{p_2}$ such as ``\textit{The answer is}'' to obtain the final answer following the question and rationale.
Formally, the output $y$ can be computed as follows:
\begin{equation}
    r \sim \prod_{i=1}^{|r|} p_{\theta}(r_i|x, \mathsf{p_1}, r_{<i}), \quad y \sim \prod_{i=1}^{|y|} p_{\theta}(y_i|x, \mathsf{p_1}, r, \mathsf{p_2}, y_{<i}).
\end{equation}

\paragraph{Few-Shot-CoT.} 
Few-Shot-CoT involves providing a set of exemplars with associated rationales. These exemplars are concatenated with the question to prompt the LLM to generate the rationale and answer.
In this setting, the output $y$ is obtained in an end-to-end mode, which can be formulated as:
\begin{equation}
    y \sim \prod_{i=1}^{|y|} p_{\theta}(y_i|E,x,y_{<i}).
\end{equation}

\subsection{Benefits of CoT}

CoT techniques have shown various kinds of benefits, including improved reasoning performance, interpretability, controllability, and flexibility. We summarize them in detail below.

\paragraph{Improved Reasoning Performance} CoT facilitates a step-by-step progression in the reasoning process for LLMs. By breaking down complex, multi-step problems into intermediate stages, CoT minimizes the risk of overlooking crucial details. Moreover, it ensures the efficient allocation of additional computational resources to problems demanding a higher degree of reasoning steps.  Numerous studies have conclusively demonstrated the efficacy of CoT across a wide range of domains, encompassing arithmetic reasoning, commonsense reasoning, and symbolic reasoning \citep{wei2023chainofthought, zero-shot,wang2023selfconsistency}.

\paragraph{Improved Interpretability} CoT offers an interpretable glimpse into the decision-making process of LLMs. Breaking down complex reasoning tasks into a chain of interconnected thoughts makes it easier to understand the underlying logic and reasoning behind a decision or conclusion made by LLM. It sheds light on how the model may have reached a specific answer, offering valuable insights for debugging and pinpointing where the reasoning process may have deviated from the correct path. However, it is important to note that fully characterizing the model's computations supporting an answer still presents an open challenge \citep{wei2023chainofthought}.

\paragraph{Improved Controllability} By prompting LLMs to output a chain of interconnected thoughts,  users can exert greater influence over the cognitive processes of LLM. Many studies \citep{yao2023tree,DBLP:journals/corr/abs-2306-03872}  were dedicated to the identification and rectification of specific thought units where the reasoning path may have gone off track or where additional information is required. This increased controllability allows for more deliberate and accurate answers.

\paragraph{Improved Flexibility} The utilization of CoT reasoning can be easily prompted in adequately large, off-the-shelf LLMs by simply adding instruction at the end of the input question for Zero-Shot-CoT or incorporating CoT exemplars used for Few-Shot-CoT \citep{wei2023chainofthought}. The flexibility of CoT extends beyond the realm of reasoning tasks, making it applicable to a wide range of fields, including classic natural language processing (NLP), scientific applications, and agent-based systems.


\section{Underlying Mechanism of CoT}
This section explores the foundational mechanisms of CoT, encompassing the general conditions that determine when and why CoT is effective.

\subsection{When CoT Works}\label{sec:when}
Although CoT has shown promising benefits, it may not be suitable in any conditions \citep{zero-shot,wei2023chainofthought,zhang2023multimodal}. We will introduce when CoT works in engineering and theoretical perspectives. Then we summarize the general conditions to suggest the effective application scopes of CoT reasoning.

$\bullet$ From an engineering perspective, \citet{wei2023chainofthought} thought that CoT reasoning is helpful under three conditions: (i) an LLM is used; (ii) the task is challenging and requires multi-step reasoning; (iii) the performance of direct prompting does not increase dramatically while scaling the model size. Notably, \citet{tay2022ul2} further provided evidence that LLMs with 20 billion parameters, pre-trained on a mixture of denoising functions, can also achieve effective CoT reasoning.\footnote{It should be noted that recent studies have explored fine-tuning smaller language models to perform CoT reasoning for specific tasks \citep{magister2022teaching,yue2023mammoth}. Here we only discuss general scenarios where LLMs can achieve effective CoT reasoning per se---better performance than direct reasoning---without additional task-specific fine-tuning on CoT-style training data.} Otherwise, CoT techniques tend to struggle with smaller LLMs \citep{wei2022emergent}. It may lead to hallucination because of lacking supportive knowledge in LLMs \citep{zhang2023multimodal} and inferior reasoning capabilities \citep{magister2022teaching}. CoT reasoning is also less effective in simple-step tasks such as matching, sequence labeling \citep{qin2023chatgpt}, and single-choice question \citep{chen2023fireact}.

$\bullet$ From a theoretical perspective, \citet{prystawski2023think} proved that CoT reasoning is helpful when training data (possibly considered as the parametric knowledge in an LLM) consists of local clusters of variables that strongly influence each other. This finding implied that the LLM must have the knowledge related to the task to support CoT reasoning. We call such knowledge as atomic knowledge.

As CoT reasoning is often elicited by in-context learning (ICL), such as Zero-Shot-CoT and Few-Shot-CoT, another line of study tries to understand when CoT works from the perspective of ICL. \citet{zhang2023automatic} showed that CoT reasoning works effectively when prompted with diverse exemplars. \citet{wang-etal-2023-towards} found that rationales being relevant to the query and correctly ordering the reasoning steps are the keys to the effectiveness of CoT prompting.

Besides prompting, introducing reasoning materials and necessary knowledge for LLMs in the training corpus has also exhibited a profound improvement in CoT reasoning ability in LLMs \citep{yu2023towards}. Recent studies found that pre-training with code data \citep{chung2022scaling} or fine-tuning (e.g., instruction tuning) with CoT-style data \citep{yue2023mammoth} is beneficial for effective CoT reasoning. That is, the CoT reasoning in the same LLMs can be improved or the CoT reasoning ability can be induced in smaller models. 

Based on the discussion above, CoT demonstrates efficacy under two overarching conditions: first, when an LLM with preferably at least 20 billion parameters is employed, and second, when the parametric knowledge within the LLM encompasses knowledge pieces that are (i) pertinent to the task at hand and (ii) maintain strong mutual interconnections.


\subsection{Why CoT Works}\label{sec:why}
Recent studies have employed both empirical and theoretical approaches in an effort to comprehend the underlying reasons for the effectiveness of CoT.

$\bullet$ Empirically, \citet{wei2023chainofthought} believed that the success of CoT reasoning constitutes a multifaceted phenomenon that likely involves various emergent abilities. Those abilities include semantic understanding, symbol mapping, topic coherence, arithmetic ability, and faithfulness. Interestingly, \citet{zhang2023automatic} found that mistakes in exemplar rationales do not lead to significant performance drops. \citet{wang-etal-2023-towards} reported a similar observation that LLMs can generate coherent reasoning steps and achieve over 80-90\% of the performance, though prompted with invalid reasoning steps in the exemplars. Those findings imply that LLMs already have an innate ability to reason after pre-training \citep{zhang2023automatic,wang2023reasoning}. CoT prompting specifies an output format that regularizes the model generation to generate step-by-step while being in order and relevant to the query \citep{wang-etal-2023-towards}. In other words, CoT techniques help \emph{compel} the model to conduct reasoning rather than teaching it \emph{how} to accomplish reasoning \citep{zhang2023automatic}. 

$\bullet$ Theoretically, Bayesian inference is a popular way to investigate why CoT works from a theoretical perspective \citep{prystawski2023think,wang2023reasoning}.
\citet{prystawski2023think} proved that CoT is effective when the training data exhibits a localized structure with respect to dependencies between variables. In the context of LLMs, the proof can be interpreted that the parametric knowledge within the LLM comprises knowledge pieces that are related to the target problem, and those knowledge pieces exert strong mutual connections with each other. To verify the proof, \citet{bi2023program} conducted an empirical study on code data and found that the local structural properties of the data are crucial for improving CoT reasoning abilities. These findings in \citet{prystawski2023think} and \citet{bi2023program} compellingly indicated that CoT may help identify the atomic pieces of knowledge used for reasoning and bridge the relationship between the atomic pieces of knowledge with intermediate reasoning steps. Similarly, \citet{wang2023reasoning} used knowledge graphs for analysis and found that organizing the known facts as ``chains'', i.e., CoT, can significantly impact the effectiveness of reasoning. By doing so, LLMs are able to accurately deduce previously unseen facts from known ones to answer a given query without explicitly encoding reasoning rules.


\section{Paradigm Shifts of CoT}\label{sec:paradigm}

\begin{figure*}[t]
    \centering
    \includegraphics[width=1.0\textwidth]{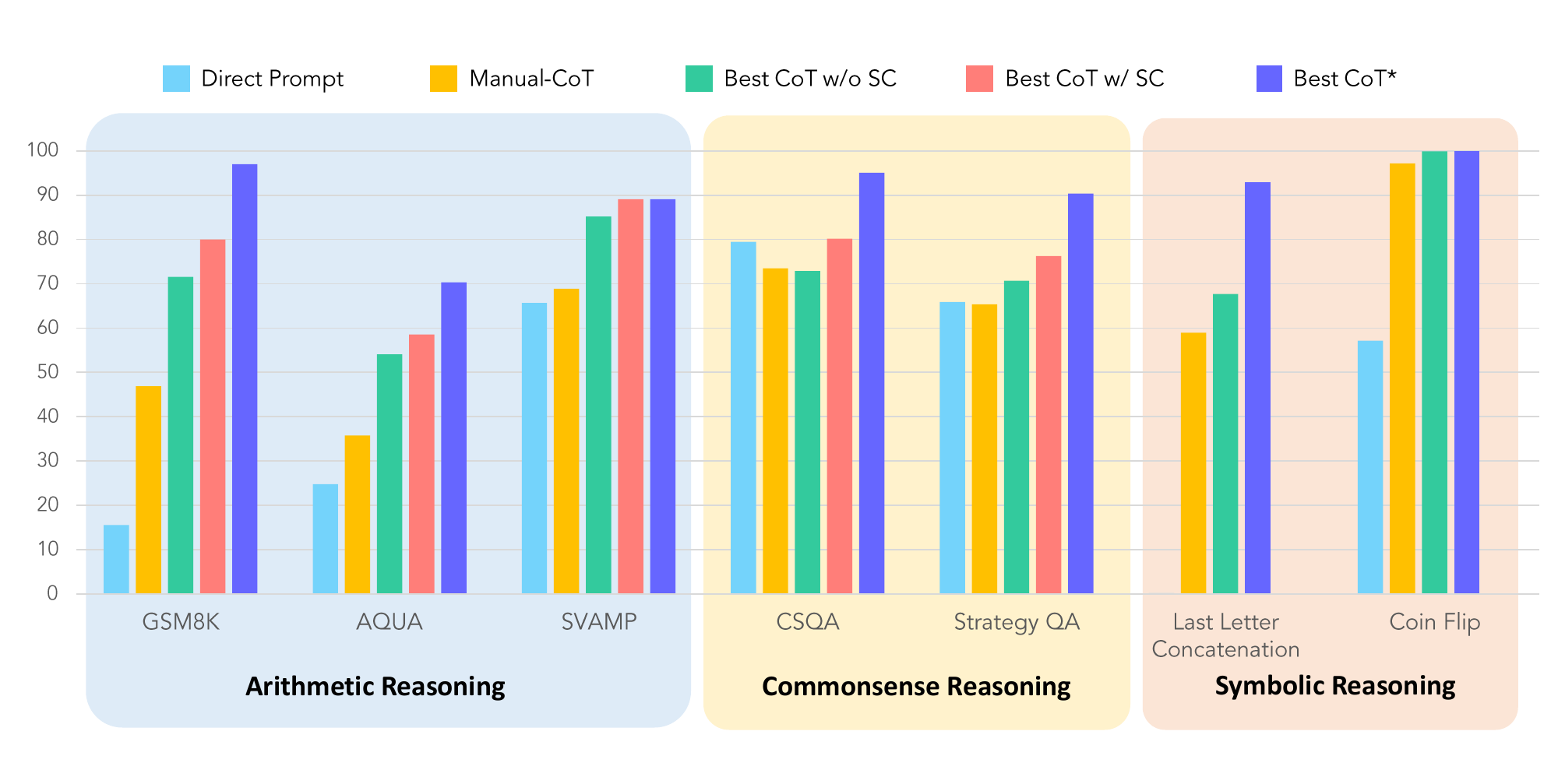}
    \vspace{-3mm}
    \caption{Performance on seven reasoning tasks. ``Direct Prompt'' refers to the standard few-shot prompting approach, where exemplars are formatted as questions and answers, with the model providing direct answers. ``Best CoT w/o SC'', ``Best CoT w/ SC'' and ``Best CoT*'' represent the highest accuracy (\%) achieved as of October 2023 (``SC'' stands for self-consistency \citep{wang2023selfconsistency}.). While the first two uses the text-davinci-002 as the LLM engine, the latter allows for the model employed to vary for each task (details in Table \ref{tab:best}).  For a fair comparison, the performances of ``Direct Prompt'', ``Manual-CoT'',  ``Best CoT w/o SC'' and ``Best CoT w/ SC'' are all based on using the text-davinci-002 as the LLM engine.}
    \label{fig:best}
\end{figure*}

After elucidating the general conditions determining when and why CoT is effective, we seek to achieve a more profound and intuitive understanding of the improvements in CoT's reasoning capabilities for LLMs. To this end, we compile and summarize the best performances of CoT across seven of the most emblematic reasoning tasks as of October 2023. We compare these performances with those achieved without CoT and present our findings in Figure \ref{fig:best}. These seven reasoning tasks span across distinct categories, including: (i) Arithmetic Reasoning: GSM8K \citep{gsm8k}, AQuA \citep{DBLP:conf/acl/LingYDB17}, and SVAMP \citep{svamp}; (ii) Commonsense Reasoning: CSQA \citep{csqa} and Strategy QA \citep{DBLP:journals/tacl/GevaKSKRB21}; (iii) Symbolic Reasoning: Last Letter Concatenation \citep{wei2023chainofthought}, and Coin Flip \citep{wei2023chainofthought}.

Figure \ref{fig:best} clearly illustrates that the benchmark performance in complex reasoning tasks has advanced rapidly, with CoT exerting a significant influence on the reasoning abilities of LLMs across all seven tasks. Notably, apart from commonsense reasoning, the relatively straightforward CoT format Manual-CoT proposed by \citet{wei2023chainofthought} substantially improves overall accuracy compared to the direct prompt in both arithmetic and symbolic reasoning.

The deficiency in CoT's performance regarding commonsense reasoning tasks has been observed both in Manual-CoT \citep{wei2023chainofthought} and Zero-Shot-CoT \citet{zero-shot}. However, when CoT is integrated with a significantly larger PaLM (540B) model, it consistently enhances commonsense reasoning. Notably, Zero-Shot-CoT also notes that the rationales generated through CoT often exhibit logical correctness or only contain human-understandable errors. This suggests that CoT encourages improved commonsense reasoning, even when the task metrics do not explicitly measure it. While Manual-CoT fails to yield performance gains in commonsense reasoning, the optimization of reasoning techniques, such as self-consistency answer aggregation \citep{wang2023selfconsistency} and automatic exemplar construction \citep{DBLP:journals/corr/abs-2302-12822}, reveals the potential of CoT to achieve remarkable results generally. 

Moreover, for ease of reference and to provide a clear overview of how CoT can achieve top performance on the seven different datasets, we have included the latest models that have achieved the best performance and the specific LLM engine they utilized in Table \ref{tab:best}.

\begin{table}[t]
\centering\small
\setlength{\tabcolsep}{2pt}
{
\caption{Best CoT* on seven reasoning tasks (SC: self-consistency by \citet{wang2023selfconsistency}).}
\label{tab:best}
\begin{tabular}{lllll}
\toprule
Category    &  Dataset  & Model                              & Best Acc & LLM                    \\ \midrule
\multirow{3}{*}{Arithmetic Reasoning}    & GSM8K                     & CSV \citep{DBLP:journals/corr/abs-2308-07921} & 97.00    & GPT-4 Code Interpreter \\
                                         & AQuA                      & Natural Program \citep{DBLP:journals/corr/abs-2306-03872}            & 70.34    & Chatgpt                \\
                                         & SVAMP                     & PoT \citep{chen2022program} + SC                            & 89.10    & Text-davinci-002       \\
                                        \midrule
\multirow{2}{*}{Commonsense   Reasoning} & CSQA                      & Manual-CoT \citep{wei2023chainofthought} + SC                      & 95.10    & PaLM 2                 \\
                                         & Strategy QA               & Manual-CoT \citep{wei2023chainofthought} + SC                      & 90.40    & PaLM 2                 \\
                                          \midrule
\multirow{2}{*}{Symbolic Reasoning}      & last letter concatenation & Natural Program \citep{DBLP:journals/corr/abs-2306-03872}                  & 92.98    & Chatgpt                \\
                                         & Coin Flip                 & Auto-CoT \citep{zhang2023automatic}                          & 99.90    & Text-davinci-002       \\ \bottomrule
\end{tabular}
}

\vspace{-3mm}
\end{table}



In conclusion, we see that compared with the vanilla prompting approach in \citet{wei2023chainofthought}, the latest CoT reasoning techniques have been strengthened throughout the full stack of the reasoning process, such as multimodal perception \citep{zhang2023multimodal,DBLP:journals/corr/abs-2305-16582,huang2023language,DBLP:journals/corr/abs-2305-02317}, automatic prompting \citep{zhang2023automatic,diao2023active}, reasoning verification \citep{DBLP:journals/corr/abs-2212-09561,DBLP:journals/corr/abs-2305-20050,DBLP:journals/corr/abs-2306-03872}, and consistency-based sampling \citep{wang2023selfconsistency,wang2022rationaleaugmented}. 

\begin{figure*}[t]
    \centering
    \includegraphics[width=1.0\textwidth]{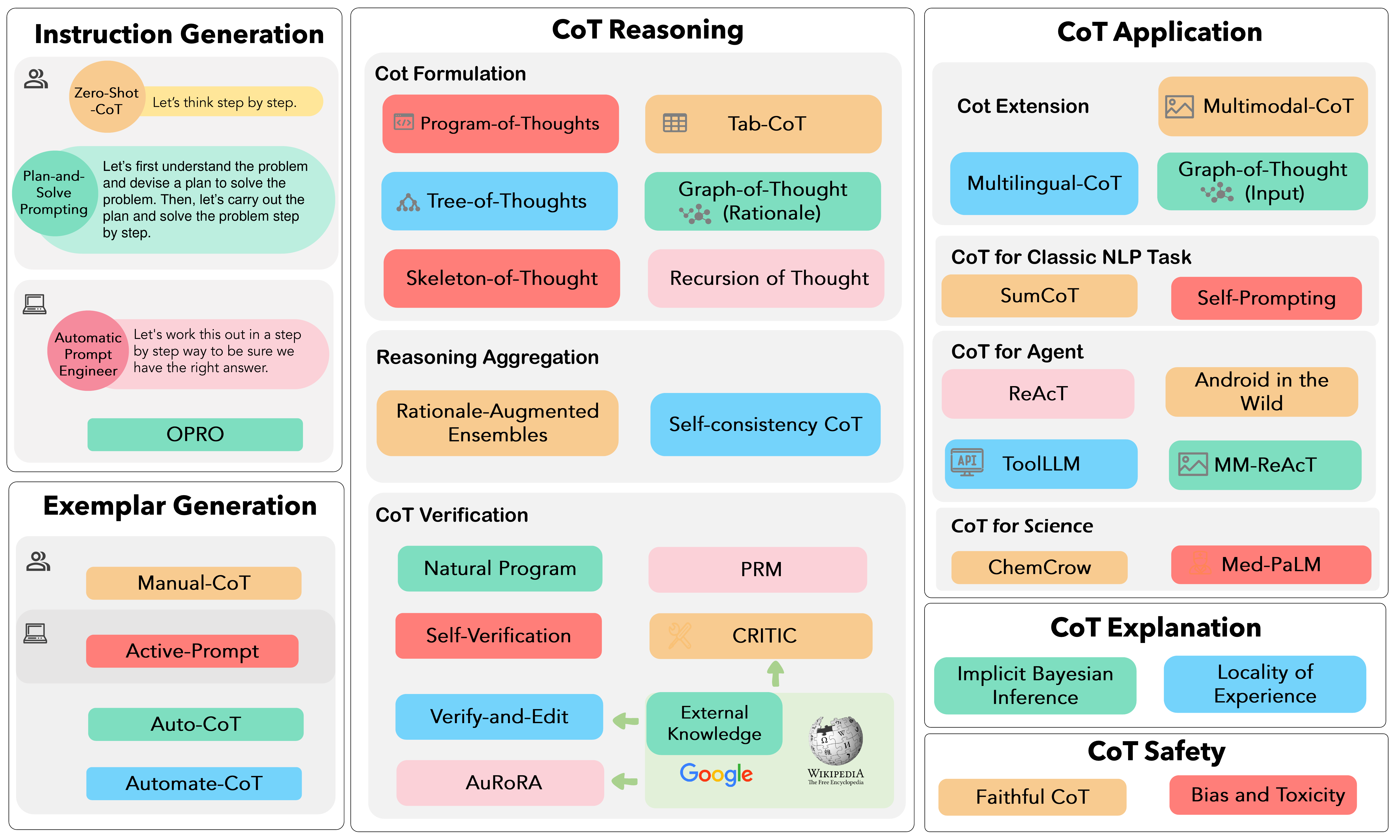}
    \caption{Overview of representative CoT approaches. We delve into the paradigm shifts of CoT techniques in three key directions: (i) \textit{prompting pattern} (instruction generation and exemplar generation); (ii) \textit{reasoning format} (CoT formulation, reasoning aggregation, and CoT verification); and (iii) \textit{application scenario} (multilingualism, multimodality, and general-purpose tasks).}
    \label{fig:CoT_family}
\end{figure*}

With the growing interest in CoT, researchers are continually striving to harness its full potential for enhancing LLMs reasoning capabilities. In this section, we will embark on a journey through the realm of CoT research, following the map of CoT overview as illustrated in Figure \ref{fig:CoT_family}, delving into the comprehensive discussions of advancements made in three key directions: (i) \textbf{prompting pattern}; (ii) \textbf{reasoning format}; and (iii) \textbf{application scenario}.

\subsection{Prompting Pattern}
The prompting pattern can be primarily divided into two components: \textbf{instruction generation} and \textbf{exemplar generation}. Instruction generation primarily focuses on finding the optimal instructions to prompt LLM, enabling them to engage in step-by-step reasoning instead of directly answering the question. This approach mainly aims to maximize LLM's zero-shot capability. Exemplar generation primarily focuses on finding the best set of input-output demonstration exemplar pairs for Few-Shot-CoT. These exemplars are used to prompt LLMs along with a test input, enabling the model to predict the corresponding output.

\subsubsection{Instruction Generation}
Instruction generation can be categorized into two distinct methods: manual instruction generation and automatic instruction generation, based on their respective generation processes.

Early efforts primarily involved manual construction of instruction prompts. The earliest and most traditional instruction generation method was Zero-Shot-CoT proposed by \citet{zero-shot}. Zero-Shot-CoT demonstrates that large language models (LLMs) can perform zero-shot reasoning by adding a simple prompt,  ``\textit{Let's think step by step}'', before each answer. Zero-Shot-CoT outperforms zero-shot LLM performances on various reasoning tasks without the need for hand-crafted few-shot examples, marking the inception of a new era in Zero-Shot-CoT.

\citet{DBLP:conf/acl/WangXLHLLL23} further proposed the Plan-and-Solve (PS) Prompting to address the missing-step errors in Zero-Shot-CoT reasoning. It consists of devising a plan to divide the task into smaller subtasks and carrying out the subtasks according to the plan. PS prompting consists of two stages. In the first stage, the author prompts the LLM using the proposed prompting template ``\textit{Let’s first understand the problem and devise a plan to solve the problem. Then, let’s carry out the plan and solve the problem step by step}'' to generate the reasoning process and the answer. The second stage extracts the answer using an answer prompt (e.g., ``\textit{Therefore, the answer (arabic numerals) is}'').

However, manually designing instructions may not always yield the desired results, and users often need to experiment with various prompts to achieve the desired behavior. In response to this challenge, \citet{zhou2022large} proposed the Automatic Prompt Engineer (APE), a method designed for the automated generation and selection of instructions for LLMs. APE treats instruction generation as a form of natural language program synthesis and optimizes this process by searching through a pool of instruction candidates proposed by an LLM. The primary goal is to maximize a chosen score function. To elaborate further,  APE initiates the process by instructing the LLM to generate a set of candidate instructions using manually crafted templates. Subsequently, it utilizes the LLM to infer the most likely instructions with the highest score, based on input-output exemplars. By harnessing the capabilities of LLMs, APE streamlines the prompt engineering process, alleviating extensive human intervention and generating high-quality instructions.

\citet{yang2023large} presented Optimization by PROmpting
(OPRO), a straightforward yet highly effective approach that harnesses the power of Language Model (LLM) as optimizers. OPRO represents a groundbreaking method in optimization, utilizing LLMs to their full potential.
OPRO initiates the optimization process by presenting a natural language description of both the optimization problem and the optimization trajectory. This trajectory includes prior solutions along with their associated optimization scores. Subsequently, updated solutions are devised and evaluated for their performance and quality. The prompt for the subsequent optimization step incorporates these solutions after thorough examination. As the iterative process unfolds, the solutions undergo progressive refinement, ultimately improving their quality.

Initially, OPRO is applied to address two classic optimization challenges: the linear regression problem and the traveling salesman problem. The study then proceeds to demonstrate that prompts optimized by OPRO surpass human-designed prompts in performance, particularly on tasks such as GSM8K and Big-Bench Hard tasks. OPRO showcases its efficiency in resolving common optimization challenges and enhancing prompts by presenting optimization tasks in natural language for LLMs, consistently generating and refining solutions.

\subsubsection{Exemplar Generation}
Similar to instruction generation, exemplar generation can also be classified into two categories based on the method of constructing exemplars: manual exemplar generation and automatic exemplar generation.

Few-Shot-CoT reasoning, formally explored by \citet{wei2023chainofthought}, represents a discrete prompt learning approach that uses multiple input-output pairs to prompt the LLM to output rationales and obtain the final answer. To provide a clearer distinction, we will refer to their work as Manual-CoT. Manual-CoT follows the traditional manual exemplar generation method. In contrast to the conventional in-context learning, where LLMs are prompted with a list of input-output demonstration pairs alongside a test input to enable the model to predict the output, Manual-CoT involves prompting the model's outputs with manually designed additional logical reasoning procedures in addition to the target output.

\citet{diao2023active} took Manual-CoT a step further by optimizing the selection of exemplars and introduced Active-Prompt, which uses task-specific example prompts annotated with human-designed rationales. Active-Prompt exists in a state that falls between manual exemplar generation and automatic exemplar generation. The method selects the most uncertain questions from a pool of task-specific queries using uncertainty-based active learning metrics.  Active-Prompt first asks LLM to answer questions multiple times following the Manual-CoT \citep{wei2023chainofthought}. The model then selects the most uncertain questions based on the uncertainty metric (e.g. disagreement, entropy, variance, self-confidence), manually annotates the rationales, and uses the questions and rationales as examples for inference. 

To eliminate the need for manual efforts in hand-crafting task-specific demonstrations 
to generate reasoning chains one by one, \citet{zhang2023automatic} proposed Auto-CoT which maintains the diversity of sampled questions and generates reasoning chains to automatically construct demonstrations. Specifically, Auto-CoT consists of two main stages: (i) Problem clustering: divide the given dataset of problems into several clusters; (ii) Demonstration sampling: select a representative problem from each cluster and use Zero-Shot-CoT to generate its reasoning chain.

\citet{DBLP:journals/corr/abs-2302-12822} proposed a strategy called Automate-CoT (Automatic Prompt Augmentation and Selection with Chain-of-Thought) that automates the process of augmenting and selecting rational chains for CoT prompting. The process consists of three steps: augmenting the language model to generate multiple pseudo-chains, pruning the pseudo-chains based on consistency with ground-truth answers, and selecting the most helpful chain-of-thought using a variance-reduced policy gradient strategy.

\subsection{Reasoning Format}
The enhancements in reasoning format primarily encompass three aspects: \textbf{CoT formulation}, \textbf{reasoning aggregation}, and \textbf{CoT verification}. {CoT formulation} focuses on transforming the sequential CoT into various cognitive structures, such as tree-like, graph-like, or table-like formats, thereby incorporating structural thinking cues. {Reasoning aggregation} primarily concerns the enhancement of LLM CoT reasoning accuracy through the aggregation of results sampled from the LLM. {CoT verification} primarily emphasizes the introduction of verification methods to verify and amend the CoT reasoning process. We will elaborate on these three aspects in the following sections.
\subsubsection{CoT Formulation}
\begin{figure*}[t]
    \centering
    \includegraphics[width=1.0\textwidth]{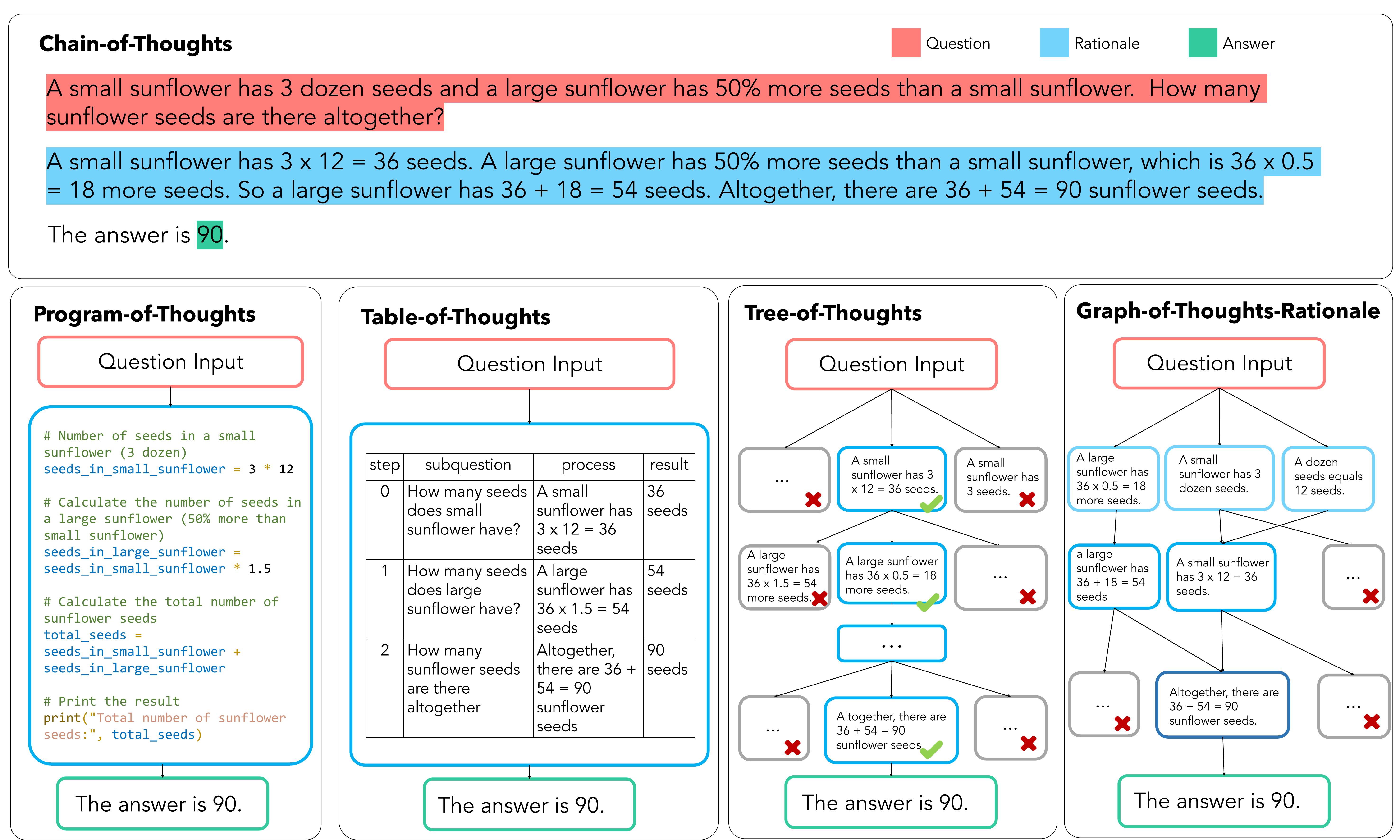}
    \caption{Formulation Shifts of CoT. We illustrate five representative CoT formulations in chronological order: (i) Chain-of-Thoughts (CoT), (ii) Programm-of-Thoughts (PoT) \citep{chen2022program}, (iii) Table-of-Thoughts (Tab-CoT) \citep{ziqi2023tab}, (iv) Tree-of-Thoughts (ToT) \citep{yao2023tree}, (v) Graph-of-Thoughts-Rationale (GoT-Rationale) \citep{DBLP:journals/corr/abs-2308-09687}. }
    \label{fig:Formulation}
\end{figure*}

We present five representative CoT formulations in Figure \ref{fig:Formulation}. We will progressively delve into the CoT formulation shifts based on this illustration.

\citet{chen2022program} introduced Program-of-Thoughts (PoT) for solving complex numerical reasoning tasks. PoT uses language models  to generate both text and programming language statements, which can be executed on a program interpreter to decouple complex computation from reasoning and language understanding.

\citet{ziqi2023tab} explored the structural reasoning ablilty for LLMs. They introduce the Tabular Chain of Thought (Tab-CoT), which adopts a table-filling approach to model CoT. In Tab-CoT, an instruction of ``$|$ step $|$ subquestion $|$ process $|$ result $|$'' is manually designed to prompt LLMs to generate a table while conducting the reasoning process. The answer is then extracted from the generated table at the end of the process. Tab-CoT showcases robust zero-shot and few-shot capabilities in performing reasoning across multiple dimensions, encompassing both rows and columns.

\citet{yao2023tree} proposed Tree-of-Thoughts (ToT) that breaks CoT into thought units and formulates them into tree structure. ToT allows LLMs to explore coherent thought units that serve as intermediate steps toward problem-solving, consider different options and evaluate their decisions. By incorporating different methods, ToT is able to look ahead to determine what to do next or trace-back to correct history decisions. Experiments have demonstrated that ToT significantly elevates the problem-solving capabilities of language models. This improvement is particularly noteworthy in the context of tasks that demand intricate non-trivial planning or search processes

Based on ToT, \citet{DBLP:journals/corr/abs-2308-09687} further extended the tree structure into graph structure and propose Graph-of-Thoughts-Rationale (To distinguish from GoT proposed by \citet{DBLP:journals/corr/abs-2305-16582}, \citet{DBLP:journals/corr/abs-2308-09687} is dubbed as GoT-rationale). 
GoT-rationale models the thought generation process of language models as a graph. The system architecture of GoT comprises multiple interacting modules: (i) Prompter that prepares prompt for the LLM, which are then used to generate responses; (ii) Parser that extracts information from the LLM's responses, which is then used by other modules in the architecture; (iii) Scorer that verifies and scores the LLM's replies to determine their quality and relevance to the task at hand; (iv) Controller that process with two elements: the Graph of Operations (GoO) and the Graph Reasoning State (GRS).
GoO is a static structure that specifies the graph decomposition of a given task, which means it prescribes the transformations to be applied to LLM thoughts, along with their order and dependencies.
GRS is a dynamic structure that maintains the state of the ongoing LLM reasoning process, which includes the history of its thoughts and their states.
Experimental results indicate that GoT outperforms state-of-the-art techniques in tasks such as sorting, set operations, keyword counting, and document merging.

\citet{lee-kim-2023-recursion} proposed Recursion of Thought (RoT), which empowers language models to recursively generate multiple contexts for problem-solving. In RoT, LLMs are prompted to output special tokens such as GO, THINK, and STOP, which serve to initiate context-related operations. The THINK token indicates the model needs to solve a sub-problem, which triggers a recursive process to generate a new context for that sub-problem.  This innovative approach enables the models to effectively handle problems whose solutions exceed the maximum context size by creating and managing multiple nested contexts.

Different from above works that focus on introducing structural information into CoT reasoning, \citet{DBLP:journals/corr/abs-2307-15337} proposed Skeleton-of-Thought to accelerate the CoT reasoning process. SoT consists of two stages: (i) Skeleton stage: SoT guides the LLM to output a concise skeleton of the answer through a manually designed a skeleton prompt template and extracts points from the skeleton response; (ii) Point-expanding stage: SoT prompts the LLM to expand on each point in parallel through a point expanding template and finally concatenates all the points to get the final answer.

\subsubsection{Reasoning Aggregation}

\citet{wang2023selfconsistency} introduced a novel decoding strategy called self-consistency to replace the greedy decoding strategy in CoT.  Self-consistency CoT first prompts the language model following the Manual-CoT \citep{wei2023chainofthought} and then samples a diverse set of reasoning paths from the language model's decoder. Finally, Self-consistency CoT finds the most consistent answer by taking a majority vote which was found to significantly improve the performance of the CoT.

\citet{wang2022rationaleaugmented} further developed a unified framework for rationale-augmented ensembles which aims at aggregating over multiple rationales generated from the language model to mitigate the brittleness of the results. The author explores three distinct approaches of rationale-augmented ensembles,each differing in how randomness is introduced into the input or output space: (i) self-consistency \citep{wang2023selfconsistency}: the ensembling is based on sampling multiple language model outputs; (ii) prompt-order ensembling: the ensembling process is based on the order of the input exemplars; (iii) input-rationale ensembling: the ensembling is based on sampling multiple input exemplars rationales from LLMs. The author found that regardless of the variation in input or prompt, the best way to improve task performance is sampling rationale in the output space.

\subsubsection{CoT Verification}
CoT verification initially focused on self-verification through multiple rounds of questioning, enabling models to validate their own responses. Later works involve leveraging external tools for information validation, such as information retrieval, calculators, or program execution. This section explores various methods and strategies within CoT verification, contributing to the enhancement of model reliability and response accuracy.

\citet{DBLP:journals/corr/abs-2212-09561} first proposed and proved that LLMs have self-verification abilities by using the conclusion obtained through CoT as a condition for verifying the original problem. Self-verification consists of two steps: (i) forward reasoning that samples multiple candidate reasoning paths; (ii) backward verification that calculates the verification scores for each candidate’s answer by masking the original conditions and predicting their results in turn. The answer with the highest score is selected as the final answer.

\citet{DBLP:journals/corr/abs-2305-20050} focused on training reward models and conducted a comparison between the outcome supervision reward model (ORM) and process supervision reward model (PRM) for LLM to solve problems from the MATH dataset \citep{DBLP:conf/nips/HendrycksBKABTS21}, finding that process supervision significantly outperforms outcome supervision. Outcome supervision is provided without humans, as the MATH dataset has automatically checkable answers. Process supervision, on the other hand, requires human data-labelers to label the correctness of each step in model-generated solutions. The authors also released PRM800K, a complete dataset of 800,000 step-level human feedback labels used to train their best reward model.

\citet{DBLP:journals/corr/abs-2306-03872} proposed a verification process using a ``Natural Program'' format.  Natural Program breaks down the reasoning process into individual steps which is accompanied by its corresponding minimal set of premises. Then, the author employed a 2-phase sequence generation, strategy Unanimity-Plurality Voting, to verify the deductive reasoning process. Unanimity-Plurality Voting first performs deductive validations on sampled reasoning chains and then conducts a majority-based voting among the verified candidate chains to obtain the final answer.

Based on Self-consistency CoT~\citep{wang2023selfconsistency}, \citet{DBLP:conf/acl/ZhaoLJQB23} designed the Verify-and-Edit framework to improve the factuality and accuracy of reasoning chains generated by CoT. The framework first passes predictions with lower-than-average consistency to the next stages for further processing. 
The second step involves producing verifying questions using manually designed prompts to test the factual correctness of the predictions. The framework then retrieves external knowledge from reliable systems (e.g., Wikipedia, Google) and edits the generated rationales with the informed answers obtained from external knowledge. Finally, the framework produces new predictions based on the edited rationales.

Similarly, \citet{gou2023critic} introduced a framework called CRITIC that allows large language models (LLMs) to validate and amend their own outputs through tool-interactive critiquing. CRITIC formulates various external tools into text-to-text functions (e.g., search engines, code interpreters) to integrate external tools into LLMs. Through a manually designed prompt template, the framework starts with an initial output and interacts with appropriate external tools to evaluate certain aspects of the text, revising the output based on the feedback obtained during the validation process.

\citet{aurora-web} proposed AuRoRA, an augmented reasoning and refining system with task-adaptive CoT prompting. AuRoRA has the characteristics of task self-adaptation and process automation. It extracts relevant knowledge from multiple sources, reducing the issue of incorrect information. Knowledge from different sources (e.g., Wikipedia) is then combined, double-checked, and refined to enhance reliability. The system revises the initial CoT using high-quality extracted knowledge to enhance accuracy and logic.

Instead of using a single LLM to refine their outputs based on feedback on their previous outputs, multi-agent debate has been proposed to improve reasoning performance \citep{du2023improving}. \citet{liang2023encouraging} identified a degeneration-of-thought (DoT) problem---the LLM fails to generate novel thoughts through reflection even if its initial stance is incorrect once the LLM has established confidence in its solutions. The DoT problem can be addressed by allowing divergent thinking using a Multi-Agent Debate (MAD) framework where multiple agents express their arguments and a judge manages the debate process to obtain a final solution. \citet{du2023improving} also leveraged multiple instances of an LLM to debate their individual reasoning processes over multiple rounds to arrive at a consistent final answer. The approach has been shown to improve the factual validity of generated content and reduce fallacious answers and hallucinations.

$\bullet$ \textbf{Can LLMs perform reliable CoT verification?} Though CoT verification approaches above have been proposed as a remedy to improve reasoning performance and reliability, the role and efficacy of the verification are questioned. Recent work has tried to examine the self-verification capabilities of LLMs in reasoning tasks \citep{valmeekam2023can,huang2023large,stechly2023gpt}. 
\citet{huang2023large} identified that the enhancements observed in CoT verification studies were often facilitated by the utilization of oracles, which guided the self-correction process using ground-truth labels, external tools, or feedback from the environment to evaluate the correctness of the responses.
However, it is crucial to note that obtaining high-quality external feedback is challenging in real-world applications. 
In the absence of oracles, LLMs encounter difficulties in rectifying their initial responses solely relying on their inherent capabilities---which we regard as \textit{imperfect verification}. In the imperfect verification scenario, LLMs tend to nonexistent violations, and over-correct the reasoning process with false positives---walk right over the correct solution especially when there are mistakes in the verification process \citep{valmeekam2023can}. This phenomenon raises concerns about the inherent capability of the LLM to accurately assess the correctness of its reasoning process. It becomes evident that the key to achieving effective CoT verification lies in harnessing external, high-quality feedback for verification. For instance, integrating external tools such as search engines and calculators into the verification process has shown beneficial \citep{chen2022program,chen2023teaching,olausson2023demystifying,pan2023automatically}.

\subsection{Application Scenarios}
Inspired by the latest techniques proposed above to enhance the reasoning capabilities of LLMs, CoT techniques have shown greater impact with the shifts of its application scenarios. The application scenario shifts include the extension from single-language tasks to \textbf{multilingual tasks}, from single-language modality to \textbf{multimodalities}, and from complex reasoning tasks to \textbf{general-purpose tasks}.

\subsubsection{From Single Language to Multilingual Scenarios} 

~\citet{shi2022language} extended the CoT to encompass the realm of multilingualism and introduces the Multilingual Grade School Math (MGSM) benchmark, which evaluates the reasoning abilities of large language models in multilingual settings. This benchmark comprises 250 grade-school math problems that have been translated into ten linguistically diverse languages. Furthermore, the authors proposed a concept called ``Multi-lingual CoT'', which involves prompting LLMs with multilingual exemplars and incorporating English intermediate reasoning steps. This approach has been shown to yield competitive or even superior results. Multi-lingual CoT suggests that employing English chain-of-thought prompting as a baseline could be a valuable strategy for multilingual reasoning research.

\subsubsection{From Text Modality to Multimodalities}

\begin{figure*}[t]
    \centering
    \includegraphics[width=1.0\textwidth]{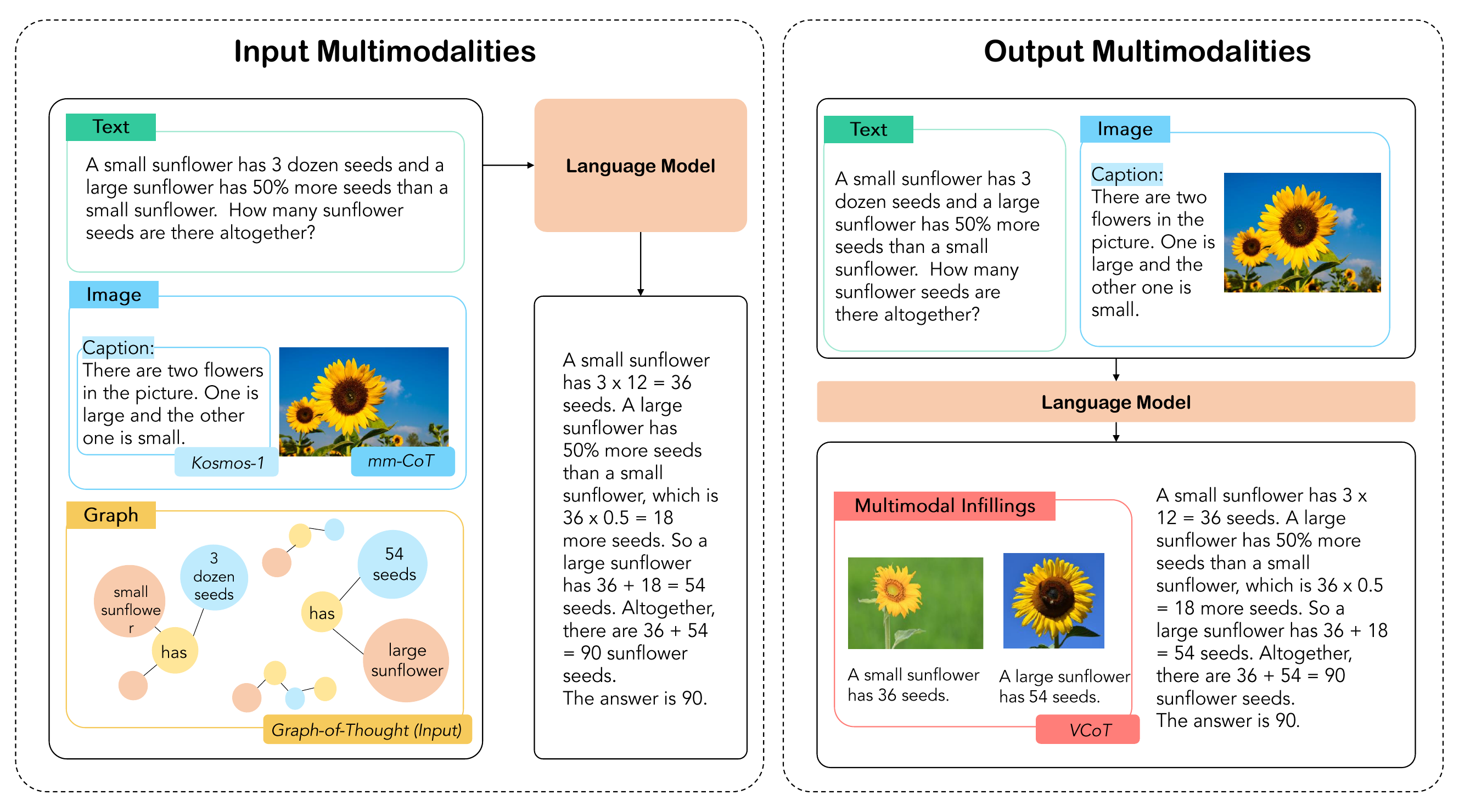}
    \caption{Formulation of multimodalities CoT. We categorized multimodalities in CoT into two types: (i) Input Multimodalities: Various modalities such as text, image \citep{zhang2023multimodal}, caption \citep{huang2023language}, and graph \citet{DBLP:journals/corr/abs-2305-16582} are incorporated into the model's input; (ii) Output Multimodalities: Multimodalities, including text and image \citep{DBLP:journals/corr/abs-2305-02317}, are introduced into the model's output.}

    \label{fig:mm-CoT}
\end{figure*}

Multimodalities in CoT can be classified into two categories: input multimodalities and output multimodalities, depending on where the multimodal elements are introduced. Figure \ref{fig:mm-CoT} illustrates these types of multimodalities in CoT.

~\citet{zhang2023multimodal} first explored input multimodalities CoT, which enables the CoT to transcend beyond textual information and proposes a multimodal CoT (MM-CoT). Instead of prompting LLMs, MM-CoT focuses on fine-tuning. MM-CoT incorporates language (text) and vision (images) modalities into a two-stage framework: rationale generation and answer inference. MM-CoT fine-tunes smaller LLMs and integrates language and visual modalities using a gated fusion mechanism. The results of this approach have demonstrated that incorporating visual information can enhance the LLM's ability to generate reasoning paths and mitigate the hallucination challenges, resulting in improved performance.

Based on \citet{zhang2023multimodal}, \citet{DBLP:journals/corr/abs-2305-16582} first introduced graph structures into input multimodalities CoT and proposes a two-stage pipeline, Graph-of-Thought-Input (GoT-Input). Different from GoT-Rationale \citep{DBLP:journals/corr/abs-2308-09687} which models the thought generation process as a graph structure, GoT-Input, on the other hand, centers its attention on modeling thought graphs derived from CoT rationales to enhance the model's reasoning capabilities.  In the first stage, the model generates the rationale given the input question and a thought graph built by leveraging open IE systems to extract the sub-verb-obj triplets from the input. In the second stage, the model generates the answer given the question and the generated rationales as inputs and a new thought graph based on the input text. GoT employs different encoders for text, graph, and image (optional) respectively and enhances the deductive reasoning capability through the usage of GNN. GoT then fuses the features using a gated fusion method to generate the final answer. By modeling the non-sequential nature of human thinking within LLMs, GoT proves to enhance the LLMs with deductive reasoning abilities

In \citet{huang2023language}, KOSMOS-1, a multimodal language model capable of processing various modalities, was introduced. The authors explored a multimodal chain-of-thought prompting approach using KOSMOS-1. In the initial stage, when presented with an image, the authors employed the prompt "\textit{Introduce this picture in detail:}" to generate a detailed description of the image as the rationale. Subsequently, the model was provided with both the rationale and a task-specific prompt to generate the final results.

In contrast to the input multimodalities CoT mentioned above, Visual Chain of Thoughts (VCoT) \citep{DBLP:journals/corr/abs-2305-02317} introduces multimodalities into the output space. VCoT initiates the process by generating captions for visual elements and identifying multipoint foveation to maintain input sequence consistency when producing multimodal infillings. Subsequently, it employs a recursive approach to generate multimodal infillings, encompassing both images and image captions. This is achieved through a combination of novelty-driven recursive infilling and consistency-driven visual augmentation. These strategies are employed to enhance interpretability for multi-step reasoning and bridge logical gaps, ultimately contributing to improved downstream task performance.

\subsubsection{From Complex Reasoning Rasks to General-Purpose Tasks}

The applicability of CoT has expanded from its initial utilization in mathematical, commonsense, and logical reasoning tasks to encompass a wide range of NLP tasks.

\citet{wang-etal-2023-element} introduced CoT into the realm of summarization and proposed the Summary Chain-of-Thought (SumCoT) technique with the aim of guiding Large Language Models (LLMs) to generate summaries in a step-by-step fashion. This approach enables the integration of more fine-grained details from source documents into the final summaries. SumCoT begins by instructing LLMs to extract core news elements from the source document using manually designed guiding question prompts. Subsequently, it involves integrating the extracted elements along with additional details from the source documents to produce comprehensive and informative summaries.

\citet{li2023prompting} proposed Self-Prompting LLMs for Open-Domain QA (ODQA). Self-Prompting consists of two stages: In the first stage, the model tasks LLM with generating a pseudo ODQA dataset by prompting it to automatically construct QA pairs with context paragraphs and explanations. In the second stage, the model dynamically selects a few examples from a pool using a clustering-based retrieval method to serve as context demonstrations. These selected examples aid in understanding and answering specific questions.

\citet{he2023exploring} explored the CoT technique in machine translation and introduced Multi-Aspect Prompting and Selection (MAPS). Drawing inspiration from strategies employed by human translators, MAPS breaks down the machine translation process into several steps. It requires the LLM to initially discern the topics and keywords of the sentence awaiting translation, and then to retrieve analogous example sentences. By integrating this extracted knowledge, the LLM produces more accurate translations.

In addition to the aforementioned classical NLP tasks, numerous studies have actively pursued the integration of CoT reasoning within the realm of science the development of automated intelligent agents.

\citet{DBLP:journals/corr/abs-2212-13138} presented MultiMedQA, a benchmark that combines six existing open question answering datasets and HealthSearchQA, a new free-response dataset of medical questions searched online. Based on the benchmark, the author then proposed instruction prompt tuning to further align Flan-PaLM to the medical domain, producing Med-PaLM. Specifically, the author used the soft prompt as an initial prefix shared across multiple medical datasets, followed by the relevant task-specific manual exemplars or instructions along with the target question. Following the CoT reasoning format, Med-PaLM’s answers to consumer medical questions compared favorably with clinician-generated answers, demonstrating the effectiveness of instruction prompt tuning. The research provides a glimpse into the opportunities and challenges of applying large language models to the medical domain.

\citet{bran2023chemcrow} incorporated CoT into the field of chemistry and proposed ChemCrow, a chemistry agent powered by LLM. Designed to tackle a wide spectrum of challenges spanning organic synthesis, drug discovery, and materials design, ChemCrow operates within the structured CoT reasoning format. Specifically, ChemCrow initially assembles a toolkit using various chemistry-related packages and software tools. The LLM in ChemCrow, guided by CoT reasoning principles, embarks on an automated and iterative chain-of-thought process. It begins by assessing the current state of the task, considering its alignment with the ultimate objective, planning the next steps and the choice of tools accordingly, and finally, solving the problem. Through the integration of 18 expert-designed tools, the LLM's performance in chemistry-related tasks is significantly improved. By integrating the CoT reasoning format, ChemCrow showcases its capacity to independently plan and execute a range of chemical syntheses, including an insect repellent, three organocatalysts, and even the discovery of a novel chromophore. This exemplifies its effectiveness in automating a diverse array of chemical tasks.

\section{Towards Language Agents}\label{sec:agents}


With improved capabilities by the advanced techniques above, CoT reasoning has yielded a broader impact on the AI community, notably fueling the development of autonomous agents in real life. Building intelligent autonomous agents that are capable of learning and acting in a distinct environment is a long-standing goal of artificial intelligence (AI) \citep{searle1969speech,wooldridge1995intelligent,maes1995agents,hendler1999there,wang2023survey,xi2023rise,zhou2023agents}. 
In light of the swift advancements detailed previously, CoT reasoning approaches have been leveraged for perception, memory, and reasoning, language agents, thereby enabling interaction within increasingly complex environments. These abilities serve as the foundation for developing autonomous agents that help solve complex tasks through human-agent and agent-agent collaboration.
 
\begin{figure*}[t]
    \centering
    \includegraphics[width=1.0\textwidth]{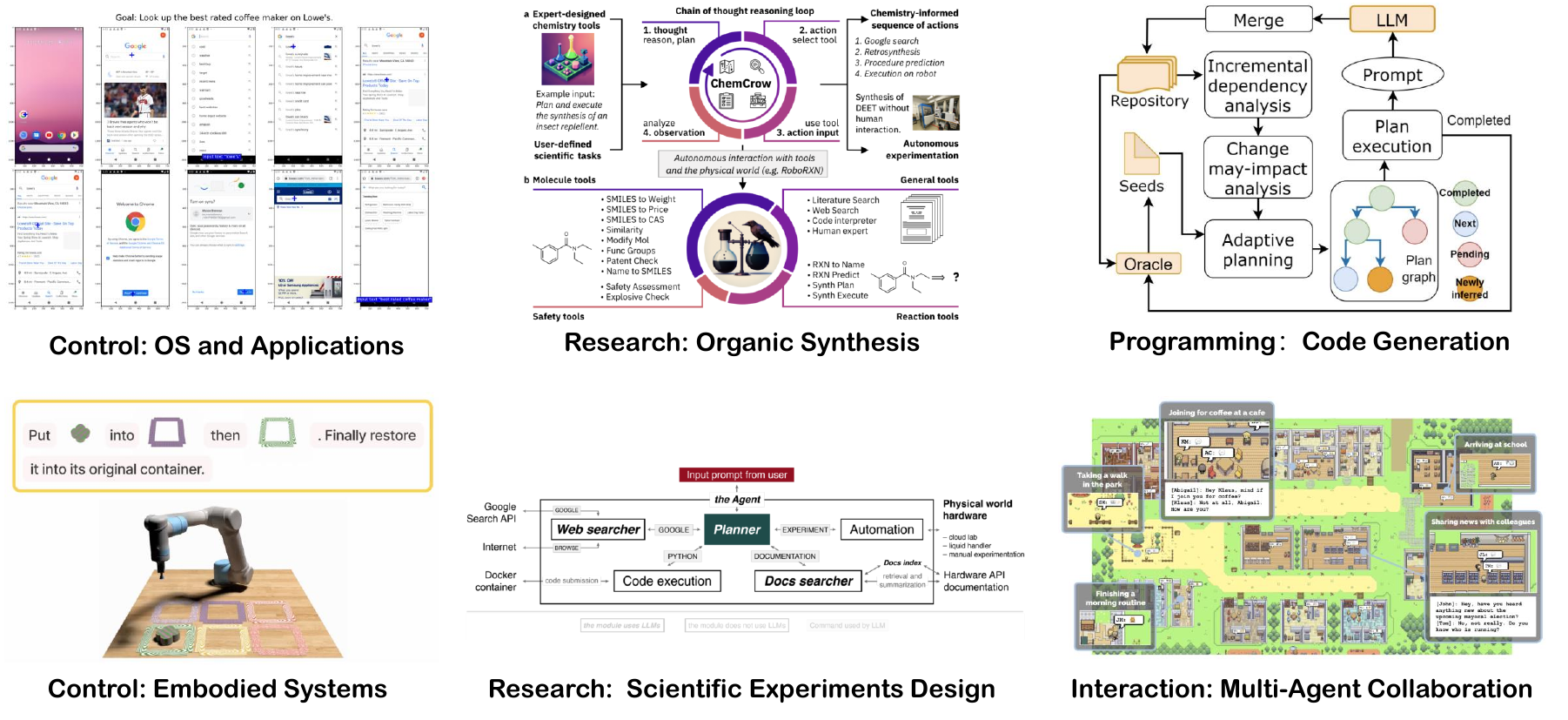}
    \vspace{-6mm}
    \caption{Representative agents for autonomous control, research, programming, and interaction. The illustrations are adapted from \citet{rawles2023android}, \citet{jiang2022vima}, \citet{bran2023chemcrow}, \citet{boiko2023emergent}, \citet{bairi2023codeplan}, and \citet{park2023generative}.}
    \label{fig:agent-example}
    \vspace{-3mm}
\end{figure*}

 As a result, LLM-based language agents, empowered by CoT techniques, have emerged in a wide range of research areas, such as engineering \citep{li2023camel, mehta2023improving, qian2023communicative}, natural sciences \citep{bran2023chemcrow, kang2023chatmof, boiko2023emergent}, and social sciences \citep{aher2023using, akata2023playing, ma2023understanding, dan2023educhat}. 
 Those language agents are capable of following language instructions and executing actions in real-world or simulated environments. 
 Figure \ref{fig:agent-example} illustrates the representative application scenarios of agents for autonomous control \citep{rawles2023android,jiang2022vima}, research \citep{bran2023chemcrow,boiko2023emergent}, programming \citep{bairi2023codeplan}, and interaction \citep{park2023generative}. A detailed technical comparison of existing agents is presented in Table \ref{tab:agent_family}. We will elaborate the technical philosophy in the following parts.

\begin{table}[htb]
\setlength{\tabcolsep}{1pt}
\centering
{
\caption{A technical comparison of representative agents. Specifically, we classify the memory modules into two main types: short-term memory and long-term memory. As defined in Section \ref{sec:memory}, short-term memory is dynamic in nature and can be easily read and written via prompts. The most common form of short-term memory is chat history. Long-term memory, on the other hand, is static and is typically stored in a database, accessible through various retrieval methods, including tree search, text search, and vector retrieval. For the external tools module, we divide the tools into three types: Web search (Web), Code interpreter (Code), and other tools (Other). More details of tool use can be found in Section \ref{sec:tool}.}
\label{tab:agent_family}
\scalebox{0.68}{
\begin{tabular}{lccccccccccc}
\toprule
\multirow{2}{*}{Agent}                                        & \multirow{2}{*}{Type}          & \multicolumn{3}{c}{Memory}                                               & \multirow{2}{*}{Methodology}        & \multirow{2}{*}{Domain}             & \multicolumn{2}{c}{Environment Interaction} & \multicolumn{3}{c}{External Tools}                           \\ \cline{3-5} \cline{8-12} 
                                                              &                                & Operation                      & Short-term         & Long-term          &                             &                                     & Modality           & Model                  & Web         & Code   & Other              \\  \midrule
CAMEL \citep{li2023camel}                                                         & Communicative                  & Prompt                         & $\checkmark$                  & -                  & Prompting                   & AI Society \& Coding                  & Text               & GPT-3.5-Turbo            & -                  & -                  & -                  \\
\midrule
Generative Agents \citep{park2023generative}                                             & Communicative                  & Tree Search                    & $\checkmark$                  & $\checkmark$                  & Prompting                   & AI Society                          & Text               & GPT-3.5-Turbo            & -                  & -                  & -                  \\
 \midrule
Voyager  \citep{wang2023voyager}                                                     & Communicative                  & Tree Search                    & $\checkmark$                  & $\checkmark$                  & Prompting                   & MineCraft              & Text               & GPT-4                  & -                  & -                  & -                  \\
 \midrule
GITM  \citep{zhu2023ghost}                                                        & Communicative                  & Tree Search                    & $\checkmark$                  & $\checkmark$                  & Prompting                   & MineCraft              & Text               & GPT-3.5-Turbo            & -                  & -                  & -                  \\
 \midrule
MetaGPT \citep{hong2023metagpt}                                                      & Communicative                  & Text Retrieval                  & $\checkmark$                  & $\checkmark$                  & Prompting                   & AI Society \& Coding                & Text               & GPT-4                  & $\checkmark$                  & $\checkmark$                  & $\checkmark$                  \\
 \midrule
ChatDev \citep{qian2023communicative}                                                      & Communicative                  & Text Summary                   & $\checkmark$                  & $\checkmark$                  & Prompting                   & Software Engineering                & Text               & GPT-3.5-Turbo            & -                  & $\checkmark$                  & -                  \\
 \midrule
MAD \citep{liang2023encouraging}                                                          & Communicative                  & Prompt                         & $\checkmark$                  & -                  & Prompting                   & Reasoning                           & Text               & GPT-3.5-Turbo          & -                  & -                  & -                  \\
 \midrule
Multiagent Debate  \citep{du2023improving}                                           & Communicative                  & Prompt                         & $\checkmark$                  & -                  & Prompting                   & Reasoning                           & Text               & GPT-3.5-Turbo          & -                  & -                  & -                  \\
 \midrule
FORD \citep{xiong2023examining}                                                         & Communicative                  & Prompt                         & $\checkmark$                  & -                  & Prompting                   & Reasoning                           & Text               & GPT-4                  & -                  & -                  & -                  \\
 \midrule
\multirow{3}{*}{AutoGPT \citep{autogpt} }                                     & \multirow{3}{*}{Autonomous}    & \multirow{3}{*}{Vector Search} & \multirow{3}{*}{$\checkmark$} & \multirow{3}{*}{$\checkmark$} & \multirow{3}{*}{Prompting}  & \multirow{3}{*}{Task Management}    & Text               & GPT-4                  & \multirow{3}{*}{$\checkmark$} & \multirow{3}{*}{$\checkmark$} & \multirow{3}{*}{$\checkmark$} \\
                                                              &                                &                                &                    &                    &                             &                                     & Speech             & DALL-e                 &                    &                    &                    \\
                                                              &                                &                                &                    &                    &                             &                                     & Image              & ElevenLabs             &                    &                    &                    \\
                                                             \midrule
BabyAGI \citep{babyagi}                                                      & Autonomous                     & Vector Search                  & $\checkmark$                  & $\checkmark$                  & Prompting                   & Task Management                     & Text               & GPT-4                  & -                  & -                  & -                  \\
 \midrule
AgentGPT \citep{agentgpt}                                                     & Autonomous                     & Vector Search                  & $\checkmark$                  & $\checkmark$                  & Prompting                   & Task Management                     & Text               & GPT-4                  & $\checkmark$                  & $\checkmark$                  & $\checkmark$                  \\
 \midrule
\multirow{2}{*}{Auto-UI \citep{zhang2023you}}                                     & \multirow{2}{*}{Autonomous}    & \multirow{2}{*}{Prompt}        & \multirow{2}{*}{$\checkmark$} & \multirow{2}{*}{-} & \multirow{2}{*}{Finetuning} & \multirow{2}{*}{UI control}         & Text               & FLAN-Alpaca            & \multirow{2}{*}{-} & \multirow{2}{*}{-} & \multirow{2}{*}{-} \\
                                                              &                                &                                &                    &                    &                             &                                     & Image              & BLIP-2                 &                    &                    &                    \\
                                                \midrule
AITW \citep{rawles2023android}                                                         & Autonomous                     & Prompt                         & $\checkmark$                  & -                  & Prompting                   & UI control                          & Text               & PaLM 2                 & -                  & -                  & -                  \\
 \midrule
DCACQ \citep{mehta2023improving} & Autonomous                     & Prompt                         & $\checkmark$                  & -                  & Finetuning                  & Engineer                            & Text               & BART                   & -                  & -                  & -                  \\
 \midrule
ChemCrow \citep{bran2023chemcrow}                                                     & Autonomous                     & Prompt                         & $\checkmark$                  & -                  & Prompting                   & Chemistry                           & Text               & GPT-4                  & $\checkmark$                  & $\checkmark$                  & $\checkmark$                  \\
 \midrule
Chatmof \citep{kang2023chatmof}                                                      & Autonomous                     & Text Search                    & -                  & $\checkmark$                  & Prompting                   & Material Sciences                   & Text               & GPT-4                  & $\checkmark$                  & $\checkmark$                  & $\checkmark$                  \\
 \midrule
IASSE \citep{boiko2023emergent}  & Autonomous                     & Vector Search                  & -                  & $\checkmark$                  & Prompting                   & Scientific Experiments              & Text               & GPT-4                  & $\checkmark$                  & $\checkmark$                  & $\checkmark$                  \\
 \midrule
TE  \citep{aher2023using}                                                          & Communicative                  & Prompt                         & $\checkmark$                  & -                  & Prompting                   & Social Science                      & Text               & GPT-4                  & -                  & -                  & -                  \\
 \midrule
CodePlan  \citep{bairi2023codeplan}                                                     & Autonomous                     & Tree Search                    & $\checkmark$                  & $\checkmark$                  & Prompting                   & Coding                              & Text               & GPT-4                  & -                  & $\checkmark$                  & -                  \\
 \midrule
\multirow{2}{*}{VIMA \citep{jiang2022vima} }                                        & \multirow{2}{*}{Communicative} & \multirow{2}{*}{Prompt}        & \multirow{2}{*}{$\checkmark$} & \multirow{2}{*}{-} & \multirow{2}{*}{Finetuning} & \multirow{2}{*}{Robot Manipulation} & Text               & T5                     & \multirow{2}{*}{-} & \multirow{2}{*}{-} & \multirow{2}{*}{-} \\
                                                              &                                &                                &                    &                    &                             &                                     & Image              & ViT                    &                    &                    &                    \\
                                                               \midrule
React \citep{yao2022react}                                                        & Communicative                  & Text Retrieval                          & $\checkmark$                  & $\checkmark$                  & Finetuning                  & Decision-Making                     & Text               & PaLM-8B                & $\checkmark$                  & -                  & -                  \\ \midrule
Reflexion \citep{shinn2023reflexion}                                                        & Communicative                  & Text Retrieval                         & $\checkmark$                  & $\checkmark$                   & Prompting                  & Decision-Making                     & Text               & GPT-4                & $\checkmark$                  & -                  & -                  \\ \midrule
ToRA \citep{gou2023tora}                                                        & Autonomous                  & Prompt                         & $\checkmark$                  & -                   & Prompting                  & Mathematical reasoning                    & Text               & GPT-4                & -                  & $\checkmark$                  & -                  \\ \midrule
Toolformer \citep{schick2023toolformer}                                                        & Autonomous                  & Text Retrieval                         & $\checkmark$                  & $\checkmark$                   & Finetuning                  & Mathematical reasoning                    & Text               & GPT-J                & $\checkmark$                  & $\checkmark$                  & $\checkmark$                  \\ \midrule
Fireact \citep{chen2023fireact}                                                        & Autonomous                  & Prompt                         & $\checkmark$                  & -                  & Finetuning                  & Question Answering                     & Text               & GPT-3.5                & $\checkmark$                   & -                  & -                  \\\bottomrule
\end{tabular}
}
}
\end{table}

$\bullet$ \textbf{What is new in language agents compared with RL agents?} The pursuit of developing generally intelligent agents has been a long-standing goal of AI research.
In the early stages, research on agents primarily RL techniques \citep{wilkins2014practical,mnih2015human}. RL agents are trained to make decisions through iterative interactions with an environment, receiving feedback in the form of rewards or penalties---correct moves are rewarded, while erroneous ones are penalized. 
This iterative process aims to minimize mistakes and maximize accurate decisions. RL agents possess a key trait: the ability to self-evolve through continuous interactions with their environments \citep{bai2023evolutionary}.
However, RL agents face limitations. They heavily rely on expert data and meticulously designed reward functions tailored for specific tasks. Consequently, their effectiveness is often confined to individual tasks, hampering their generalization capabilities to novel tasks or domains \citep{kim2023language}. Furthermore, the inner workings of RL agents often lack transparency and interpretability \citep{lundberg2017unified,yang2018learn}.
In contrast, language agents distinguish themselves from RL agents by leveraging commonsense priors embedded in LLMs. These priors reduce dependence on human annotation and trial-and-error learning, enabling easy adaptation to new tasks or environments and allowing better interpretability with CoT \citep{yao2022react,shah2023navigation}. However, language agents face challenges in evolving their parameters in response to environmental changes, primarily because they are predominantly adapted to environments through prompts or the heavy costs of fine-tuning the LLMs. While recent studies on language agents, such as Retroformer \citep{yao2023retroformer}, have incorporated RL-like policies to enhance the capabilities of language agents, the focus remains largely limited to language reasoning tasks.
It holds promise to see how to bridge the gap between RL agents and language agents to facilitate future architectures that can work generally with strong performance and high interpretability in complex environments. In consideration of the pros and cons of RL agents and Language agents, please refer to Table \ref{tab:RL_agent_VS_Language_agent} for more details.

\begin{table}[!htb]
\centering
\caption{Comparison between RL agents and language agents.}
\scalebox{0.68}{
\begin{tabular}{lll}
\toprule
\multicolumn{1}{c}{Aspect}                 & \multicolumn{1}{c}{RL Agents}                                                                                                               & \multicolumn{1}{c}{Language Agents  }                                                                                                                                          \\ \midrule
Knowledge Aquisition & Primarily use RL techniques.                                                                 & Leverage   commonsense priors embedded in LLMs.                                                                                     \\\midrule
Training Process     & \makecell[l]{Trained through iterative interactions with the environment, \\ receiving rewards or penalties.}                        & \makecell[l]{Adaptation to new tasks or environments with  reduced dependence on \\ human annotation,  primarily through prompts.}                                   \\\midrule
Self-Evolution         & \makecell[l]{Possess the ability to self-evolve through \\ continuous interactions with the   environment. }                           &  \makecell[l]{Face   challenges in evolving parameters in response to environmental changes; \\   Adaptation is mainly through prompts or costly fine-tuning of LLMs.}  \\\midrule
Limitations            & \makecell[l]{ Heavily   relies on expert data and task-specific reward functions;\\ Effectiveness often   confined to individual tasks.} &  \makecell[l]{Challenges   in evolving parameters dynamically; \\ Focus on language reasoning tasks, may   lack adaptability to broader tasks.}                         \\\midrule
Transparency           & \makecell[l]{Working mechanism often lacks transparency and interpretability. }                                                         & \makecell[l]{Generally allow better interpretability with commonsense priors, \\ but have challenges in parameter evolution transparency.}                            \\\midrule
Generalization         & \makecell[l]{Limited  generalization capabilities to novel tasks or domains.}                                                        &  \makecell[l]{Facilitates easy adaptation to new tasks or environments,  \\ reducing dependence on task-specific training. \\ Primary focus remains on language tasks.} \\\midrule
Future Goals           & \multicolumn{2}{l}{ \makecell[l]{Aim  to bridge the gap between RL agents and language agents, facilitating more versatile and adaptable architectures.}}                                                                                                                                   \\ \bottomrule
\end{tabular}
}

\label{tab:RL_agent_VS_Language_agent}
\end{table}

The following part will introduce the basic concepts of language agents and show how CoT is utilized in those agents.

\subsection{General Framework}

The landscape of language agent frameworks within the existing literature is notably diverse. We outline representative architectures in recent studies and summarize a cohesive and overarching conceptual framework for language agents.

\citet{wang2023survey} designed a modulized agent framework with four modules: (i) a profiling module to identify the role of the agent, (ii) a
memory module to recall past behaviors, (iii) a planning module to plan future action, and (iv) an action module to translate the agent’s decisions into specific outputs. 

\citet{xi2023rise} proposed a conceptual agent framework with three components: (i) brain that undertakes basic tasks like memorizing, thinking, and decision-making, (ii) perception that perceives and processes multimodal information from the external environment, and (iii) action that carries out the execution using tools and influences the surroundings.

\citet{zhou2023agents} presented a featurized agent framework for language agents, which supports important features, including planning, memory, tool use, multi-agent communication, and fine-grained symbolic control.

\citet{sumers2023cognitive} proposed another conceptual architecture for language agents called CoALA. CoALA organizes agents along three
key dimensions: (i) information storage that is divided into working and long-term memories, (ii) action
space that is divided into internal and external actions, and (iii) decision-making procedure that is structured
as an interactive loop with planning and execution.

Though different architectures have been designed, recent technical research \citep{yao2022react,yao2023retroformer,park2023generative,zhu2023ghost} tends to follow the line of the conceptual framework by prompting LLMs to imitate the agent processes such as perception, memory, and reasoning. The basic assumption is that LLMs have already captured world knowledge to some extent \citep{gurnee2023language}, which can be induced by CoT prompting step by step.

\begin{figure*}[t]
    \centering
    \includegraphics[width=1.0\textwidth]{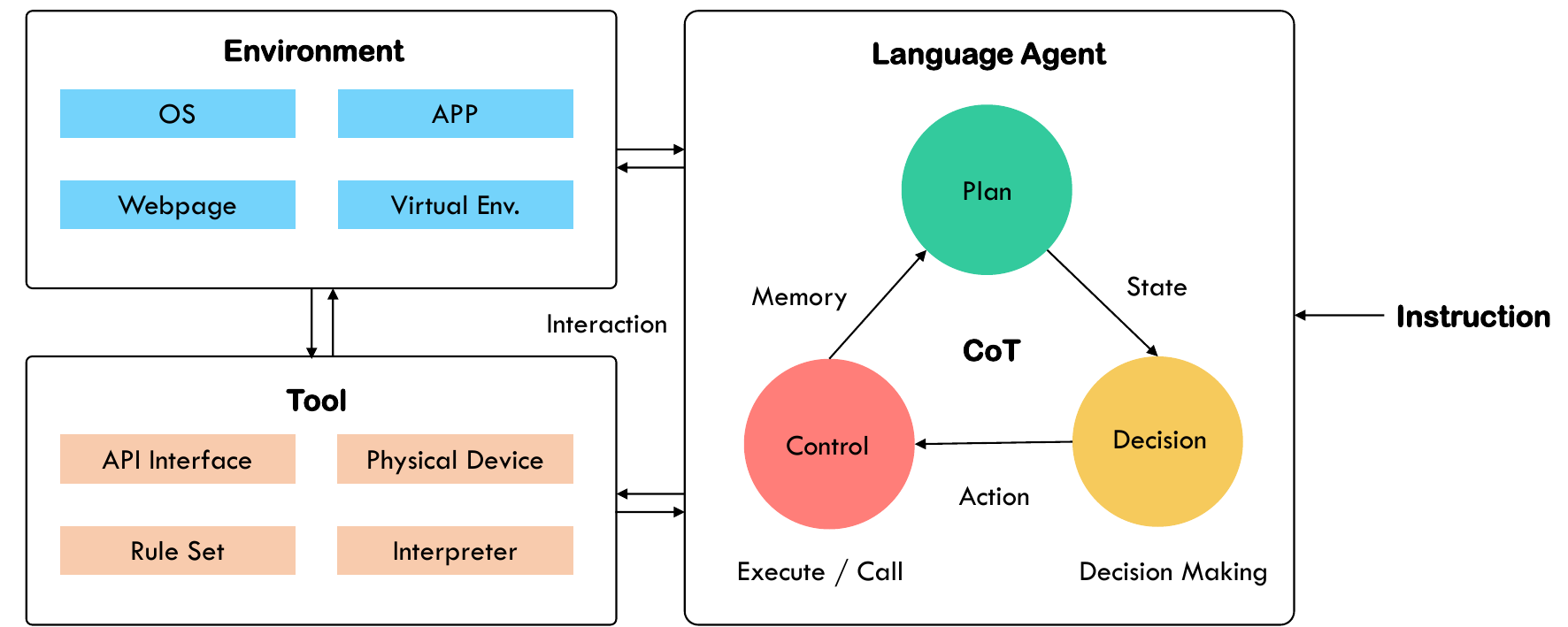}
    \caption{General framework of language agents. Language agents are capable of following language instructions and executing actions in real-world or simulated environments.}
    \label{fig:agents}
\end{figure*}

Therefore, we summarize a general conceptual framework of language agents in view of technical practice as shown in Figure \ref{fig:agents}. Given a user instruction (also known as a \textit{goal}), an agent needs to complete the task with multiple steps of interaction across the environment, possibly operating with tools. Without the loss of generality, we focus on a single agent when introducing the framework. It is worth noting that multiple agents can cooperate or compete with each other in a multi-agent environment. Before diving into the technical discussion, we first present the basic concepts of language agents, i.e., agent, environment, and tool use, as below.

\subsubsection{Agent Backbone Model}
A language agent can be built upon either a single-modality LLM or a multimodal LLM. Completing a task often comes with multiple steps of interaction. The entire process is called an \textit{episode}, which is composed of a series of \textit{turns}. To accomplish the task, the agent needs to plan ahead, make decisions, and execute actions at each turn of the episode. The process of planning, decision-making, and action execution may reflect the reasoning ability of LLMs as LLMs are exposed to real-world or virtual environments that do not exist during the pre-training of LLMs. In such environments, the LLM must perceive the world's knowledge and take action, in which cases we will show that CoT helps bridge the gap between the environment perception and the innate ability of LLMs.

Such agents expand the landscape of language models to compete in specific fields, including application operation, web searching, and web shopping. There are two popular types of language agents: autonomous agents and communicative agents. 
Typical examples of autonomous agents are AutoGPT \citep{autogpt}, BabyAGI \citep{babyagi}, and AgentGPT \citep{agentgpt}. In contrast, communicative agents are personalized and socialized agents with human behaviors that can communicate \citep{park2023generative,wang2023voyager,zhu2023ghost}, collaborate \citep{hong2023metagpt,qian2023communicative} and debate \citep{liang2023encouraging,du2023improving,xiong2023examining} with each other. They are often deployed in immersive environments. 

\subsubsection{Environment Interaction}
An intrinsic characteristic of language agents is communicating, interacting, and evolving with environments. Such environments include operation systems, third-party applications, webpages, and virtual environments. LLMs handle environments with two kinds of approaches, namely, \textbf{environment parsing} and \textbf{multimodal perception}, depending on whether the LLM has the ability to model the multimodal inputs. Environment parsing refers to those approaches that leverage external tools such as optical character recognition (OCR) and icon detectors \citep{zhang2021screen,sunkara2022towards} to parse the environment into textual elements (e.g., HTML layouts) as inputs to an LLM. In contrast, multimodal perception, also dubbed as first principles thinking \citep{zhang2023you}, refers to using a multimodal LLM to simultaneously process the inputs in different modalities. To build a multimodal LLM, a popular way is to use a simple projection matrix to integrate a pre-trained large vision model (e.g., CLIP \citep{radford2021learning} and BLIP-2 \citep{li2023blip}) into an LLM \citep{liu2023visual,zhang2023llama}. More recent studies have also explored modeling the inputs of different modalities into the same vector space, thus resulting in any-to-any representation learning \citep{huang2023language,wu2023next,moon2023anymal} and interleaved multimodal representation learning \citep{li2023textbind,zhao2023mmicl}.





\subsubsection{Tool Use} \label{sec:tool}

Tool use can be seen as an expansion of a language model's ability boundary, compensating for parametric knowledge for reasoning and grounding the language model's capabilities to interact with environments \citep{qin2023toolllm}. Tools coming into play include knowledge bases, search engines, code interpreters, online models, applications, databases, and even bespoke tools specially created for specific tasks, overcoming the constraints of generic APIs \citep{li2023api, schick2023toolformer,cai2023large,zhou2023agents,xagent2023}.


The purpose of tool use comes with three aspects: 

$\bullet$ \textbf{Action execution.}
The language model is not confined to merely predicting the next action; it has the capability to execute it in the real environment. This includes everything from executing codes or queries through a JavaScript element selection on a webpage \citep{zhou2023webarena}, executing programs via code interpreters or compilers \citep{gur2023real, ni2023lever, surismenon2023vipergpt, ruan2023tptu, gou2023tora}, to interacting with online expert models which serve as callable APIs \citep{shen2023hugginggpt, patil2023gorilla, ge2023openagi}. These steps can be dynamically adjusted with effective scaling of the tool set depending on task requirements and computational capacity \citep{Yuan2023CRAFTCL}.

$\bullet$ \textbf{External knowledge acquisition.}
Retrieval augmentation has been shown so effective that has been regarded as a standard solution to alleviate the factuality drawback \citep{trivedi2022interleaving, yao2022react}. To empower the CoT process, up-to-date knowledge is accessible through search engines \citep{khattab2022demonstrate, nakano2021webgpt}, while domain-specific through expert candidates \citep{bran2023chemcrow, ge2023openagi}.
The purpose of tool use extends beyond augmenting the language model's scope; they enable language models to adapt to a complex environment or a vast application ecosystem and ensure that the information language models have access to is up-to-date, thereby reducing the propensity to generate non-factual information \citep{wang2023survey}.


$\bullet$ \textbf{Reasoning and verification.} 
In the reasoning process, language models are sometimes prone to errors. Tools that provide accurate, real-time knowledge can help correct reasoning errors and formulate more accurate responses. Pieces of evidence from these tools are used to rewrite the initial output for self-correction \citep{gou2023critic}. Code LLMs can be further verified with execution results from program executors \citep{ni2023lever}. Multi-tool and multi-step planning and retrieval strategies, involving depth-first or breadth-first approaches, can be deployed for a deep or diverse range of possible pathways \citep{liu2023controlllm,qin2023toolllm}.


\subsection{CoT Facilitates Agent Abilities}\label{sec:cot-role}
Language agents are placed in interactive loops with the external environment \citep{sumers2023cognitive}. The interface loops can be elicited in three ways (Figure \ref{fig:cot-role}), namely, perception, memory, and reasoning. 
CoT methods empower the agents from all three perspectives.

\subsubsection{Perception as CoT} Prompting the agent to interpret the perception step by step, as a chain of perception, has been shown to improve the action success rate. It enhances the understanding of the environment or the context. Notably, \citet{rawles2023android} found that using the CoT template, \textit{``Answer: Let’s think step by step. I see <Screen Caption>, I need to ...''}, substantially improves the action prediction accuracy. As an example shown in Figure \ref{fig:cot-role}, the prompt of perception as CoT can be \textit{``Let’s think step by step. I see unrelated search results in the Google app''}. Furthermore, \citet{zhang2023multimodal} and \citet{huang2023language} leveraged external tools to obtain the image captions as supplemental inputs to help improve the perception of the multimodal environments. The captions are placed in <Screen Caption> to organize the input prompt.

In addition to the one-way interpretation of perception, language agents can benefit significantly from integrating environmental feedback, especially in the context of multi-turn interactions where the environment is subject to alterations \citep{chen2023teaching,olausson2023demystifying,jignasu2023towards}. Effectively integrating this feedback necessitates the implementation of a crucial method: self-correction with environment feedback \citep{xu2023lemur,zhou2023language,yao2023retroformer,zhao2023large}. Self-correction entails exposing the model to intricate sequences of operations, encompassing tasks such as executing codes, conducting operations, and controlling robots. These operations can lead to execution failures and generate error messages. In this context, the agent is not only required to comprehend these environmental cues but must also actively engage in iterative error correction processes until the desired outcome is achieved. Consequently, the agent's performance within these dynamic environments serves as a direct indicator of its self-correction proficiency. This proficiency, in turn, showcases the agent's ability to assimilate feedback from the environment effectively. The seamless incorporation of such feedback not only refines the interpretive capacities but also enhances its overall functionality, making it pivotal in the realm of advanced language agents.

\begin{figure}[t]
    \centering
    \includegraphics[width=1\linewidth]{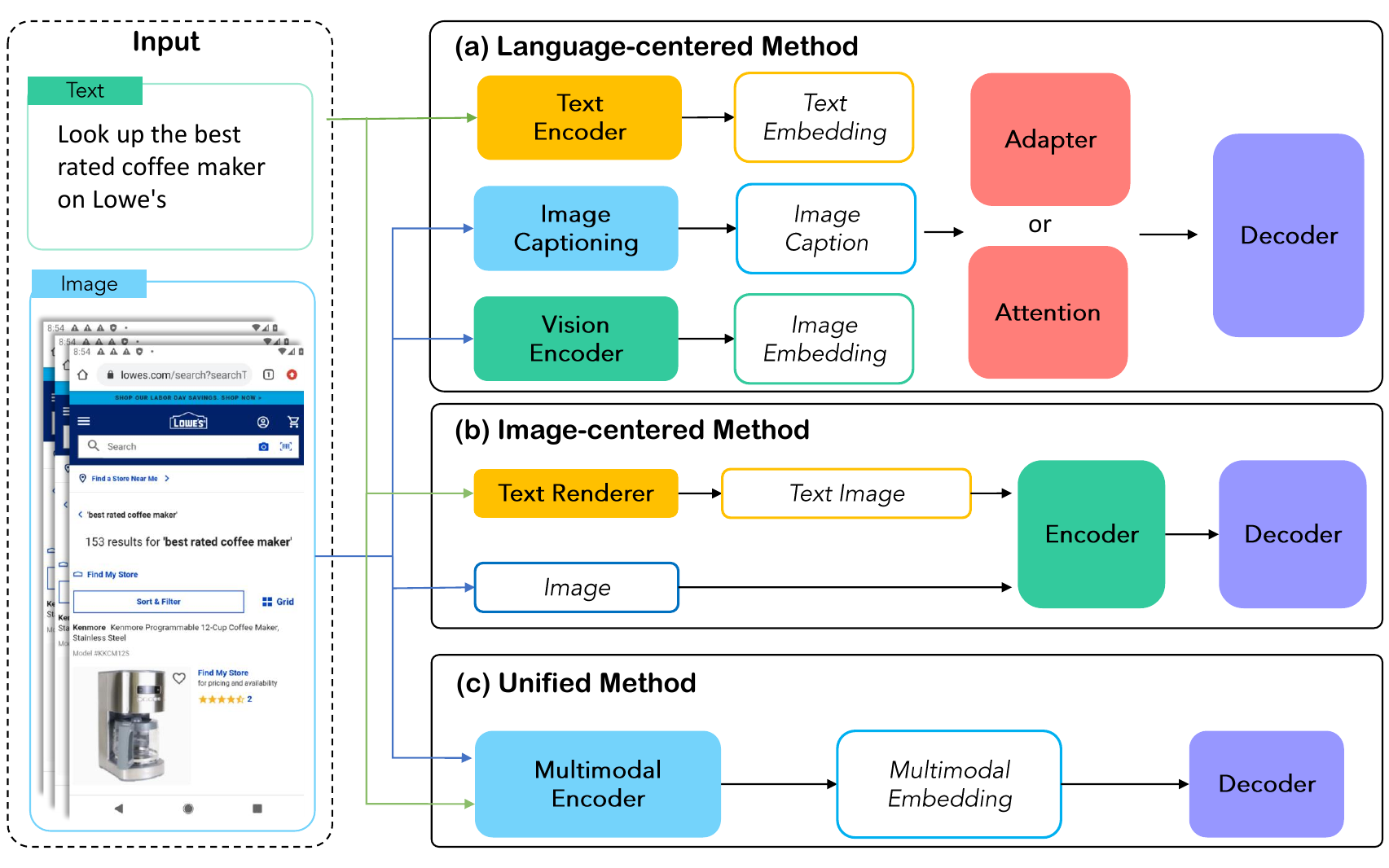}
    \vspace{-3mm}
    \caption{Multimodal perception methods including (a) language-centered method; (b) image-centered method; (c) unified method.}
    \vspace{-3mm}
    \label{fig:perception_approaches}
\end{figure}

$\bullet$ \textbf{Is language-centered perception the future?} Multimodal
perception stands as one of the key steps toward achieving artificial general intelligence. 
Current trends, likely inspired by the impressive reasoning capacities of language models, predominantly adopt a language-centered perception approach (Figure \ref{fig:perception_approaches}(a)).
Typically, distinct encoders are utilized to process inputs from various modalities, such as images. The resulting encodings are then linked to an existing language model through cross-attention or supplementary adapters, facilitating the integration of multimodal inputs into the language model's embedding space \citep{alayrac2022flamingo,liu2023improved,wu2023next,driess2023palm,chen2023pali,bai2023qwen,zhang2023llama}. 
In contrast to this prevailing language-centric modeling, \citet{rust2023language} has proposed an image-centered approach (Figure \ref{fig:perception_approaches}(b)) by rendering text as images, enabling the transfer of representations across languages based on orthographic similarity or the co-activation of pixels. 
To better align the inputs from different modalities and allow for convenient scaling up model parameters, recent research endeavors have explored a unified approach (Figure \ref{fig:perception_approaches}(c)). 
For instance, in the context of vision-language modalities, instead of employing a separate image encoder, image patches are treated as tokens and linearly projected into the embedding layer of the transformer. These patches are then fused with the representations of language tokens, allowing for seamless integration \citep{huang2023language,fuyu-8b}. 

Though various kinds of perception approaches, including language-centered, image-centered, and unified methods, have been proposed in the realm of agent perception, determining the most suitable choice remains a formidable challenge. 
This difficulty arises due to the involvement of more diverse and complex modalities such as auditory, tactile, and brain signals during interactions between agents and environments. 
Besides, these modalities often come with imbalanced data scales, complicating the perception process. 
Additionally, the diversity in types and formats of multimodal data poses challenges related to computation efficiency and the scalability of models. 
Exploring innovative methods to address these challenges will pave the way for the development of effective and efficient perception frameworks in the future.


\subsubsection{Memory as CoT} \label{sec:memory}



A language agent is commonly equipped with both long-term memory and short-term memory \citep{sumers2023cognitive,wang2023large}. 

\paragraph{Short-term memory.} Short-term memory is formed as temporal information that may be flexible to change in different steps of episodes (also known as \textit{working memory} in \citet{sumers2023cognitive}. Short-term memory is more temporal-specific, offering explicit, recent context that facilitates the agent. On the one hand, short-term memory shows direct support and closer relations with the exact current state. On the other hand, short-term memory yields a relatively moderate impact on the whole environment. For example, short-term memory can be modeled within an episode of a multi-step task, the chain of action history \citep{zhang2023you}, or the rationales or sub-question in the last several hops of multi-hop question answering \citep{yao2022react, khattab2022demonstrate}. 
Due to the significant temporal character, short-term memory raises little storage concern.

\paragraph{Long-term memory.} Long-term memory provides the agent with the capability to retain and recall static information over episodes \citep{weng2023prompt}.
In contrast to short-term memory, long-term memory is more general to the task, as a macroscopic and abstract understanding of the whole world. This can include \textit{procedural memory} that stores the production system itself, \textit{semantic memory} that stores facts about the world, and \textit{episodic memory} that stores sequences of the agent’s past behavior \citep{sumers2023cognitive}.
For example, given a goal, \textit{upvote the latest post}, in the varied environment states, two chains of actions have been observed to accomplish the goal: (i) [\textit{opening Instagram, going to home feed, looking at the latest post, upvoting the latest post}] and (ii) = [\textit{go to the HOME screen, opening Instagram, going to home feed, looking at a post, upvoting the latest post}]. It can be found that atom actions [opening Instagram, going to home feed, looking at the latest post, upvoting the latest post] can serve as long-term memory for this goal, i.e., a chain of static memory.

Long-term memories can rely on both parametric and non-parametric knowledge storage. They can be from the trainable parameters of the language agents or maintained as external knowledge that can be leveraged through retrieval systems.
For example, the earlier hops of former episodes are long-term memories from agent parameters, and the output action formulations are parametric long-term memories.

$\bullet$ \textbf{Towards efficient memory operation.} Modeling memory as linear natural language sequences becomes inefficient as sequences lengthen during the agent's interaction with environments. Besides, the context window of LLMs is predetermined to be limited in length. To pursue more efficient memory operations, recent studies have explored two types of approaches, i.e., leveraging (i) tree search and (ii) vector retrieval. 

(i) \textbf{Tree search.} Memory can be stored with a tree structure and fetched by searching on the tree. Notably, MemWalker \citep{chen2023walking} empowered agents to access textual memory information through iterative prompting. In this approach, the agent initially processes the lengthy context into a tree of summary nodes. Upon receiving a query, the agent navigates this tree to search for relevant information and responds after gathering sufficient information. Similarly, GITM \citep{zhu2023ghost} proposed an LLM Decomposer that recursively decomposes goals into a sub-goal tree. The hierarchical tree structure helps the model to explicitly capture the relationships between goals and corresponding plans in the memory. \citet{park2023generative} proposed the reflection tree to organise the memory of a communicative agent. When facing the trivial observations during the interaction with the environment, the agent periodically reflects on existing memories in a abstract manner, thus forming a reflection tree: ``the leaf nodes of the tree represent the base observations, and the non-leaf nodes represent thoughts that become more abstract and higher-level the higher up the tree they are''.

(ii) \textbf{Vector retrieval.} The other way to store memory is via vector storage \citep{hu2023chatdb,zhou2023llm}. Vector database has become a key carrier for storing, managing, and retrieving high-dimensional data, such as the long-term memory of language agents. It can represent complex data types such as text, images, videos, and even structured data. Agentsims \citep{lin2023agentsims} employed a vector database to enable efficient storage and retrieval within long-term memory. Specifically, it stores daily memories as embeddings within this vector database. When the agent encounters new situations and necessitates the recall of past memories, the long-term memory system adeptly retrieves pertinent information, thereby ensuring the consistency of the agent's behavior.

\subsubsection{Reasoning as CoT} Inspired by the success of eliciting LLMs' step-by-step reasoning abilities, CoT has also been applied in inducing the agents to reason via planning or decision-making.
More importantly, CoT methods for language agents require careful design to handle the action execution and state observation. 

The gap between reasoning and action is bridged by combining interleaving thought, action, and observation \citep{yao2022react, khattab2022demonstrate, shinn2023reflexion}.
By exploring the use of LLMs to generate both CoT traces and task-specific actions in an interleaved manner, it has been found that reasoning and acting achieve mutual promotion. Reasoning traces help the model make action plans and handle exceptions, while actions allow the LLM to interface with external sources, such as knowledge bases or environments, to gather additional information for knowledge support. 
\citep{xu2023rewoo} detached the reasoning process from external observations to reduce token consumption during multiple steps of CoT.

Similarly, AgentBench \citep{liu2023agentbench} compelled language agents to complete tasks via ``think'' and ``Act'' steps. Further, \citet{zhang2023you} proposed a chain-of-action technique---leveraging a series of intermediate previous action histories and future action plans---to help the agent decide what action to execute, which transforms the decision-making as a CoT reasoning problem. 

$\bullet$ \textbf{How to expand the capability of agents?} 
Currently, the mainstream interest is to apply CoT prompting approaches to elicit LLMs' reasoning abilities during the interaction with the environments as discussed above. 
The basic hypothesis is that LLMs already have the prior knowledge to perform as the language agents for our concerned tasks and CoT prompting approaches are effective in invoking the knowledge. 
Those prompting techniques have the advantage of flexibility and convenience because it is easy to design and adjust the prompts according to the task requirements and characteristics. 
However, LLM performance has shown to be sensitive to prompts and there is a lack of evidence that LLM can actually learn domain knowledge from the prompts. 
Therefore, purely prompting methods may not be adequate to make LLMs generalizable to new domains. 
To expand the capability boundary of language agents, there is a recent interest in fine-tuning LLMs on curated datasets to build effective agents.
\citet{chen2023fireact} called for a re-thinking of fine-tuning language models when the target tasks and data formats are known and enough data can be collected (e.g., possibly automatically with GPT-4). 
The results have revealed that fine-tuning can not only achieve strong generalization and robustness but also improve performance. 
\citet{gou2023tora} curated interleaved tool-use data composed of natural language CoT with tool-integrated programs. Then, a tool-integrated reasoning agent was trained on those high-quality annotations and achieved substantial performance gains on various mathematical reasoning tasks.


\section{Challenges}\label{sec:challenges}
Despite the swift advancements in the realms of LLMs, CoT reasoning, and language agents, numerous promising challenges still beckon for deeper exploration, particularly pertaining to generalization to unseen domains, enhancing efficiency amidst redundant interactions, developing customizable agents, scaling up language agents, ensuring the safety of language agents, and capacity evaluation.

\subsection{Generalization to Unseen Domains}
Language agents have found extensive applications in practical fields such as engineering \citep{li2023camel, mehta2023improving, qian2023communicative}, natural sciences \citep{bran2023chemcrow, kang2023chatmof, boiko2023emergent}, and social sciences \citep{aher2023using, akata2023playing, ma2023understanding, dan2023educhat}. Despite their widespread use, a significant challenge persists: adapting LLMs to specific, especially unseen domains. This challenge is twofold: firstly, determining an efficient method for acquiring domain-specific knowledge, such as employing CoT prompting techniques. The limitations arise from the finite scope of knowledge acquisition during pre-training on textual corpora, lacking substantial interaction with the physical world. Secondly, there is the challenge of effectively adapting LLMs to diverse, unseen domains. Given the substantial variation in action spaces across tasks (e.g., drone control versus web browsing), aligning the model's knowledge with the specific task requirements remains a formidable obstacle. These challenges underscore a critical gap in current research. The need to enhance LLMs' adaptability to novel domains and help LLMs learn from environments is paramount, requiring innovative solutions that address both knowledge acquisition and effective task alignment. 

Prompting and fine-tuning are widely used techniques to adapt pre-trained LLMs to new domains. However, it remains an underexplored area of when and how to leverage prompting (e.g., prompting pattern and reasoning format) and fine-tuning (e.g., instruction tuning) techniques to help LLMs generalize to unseen domains.
In doing so, researchers can pave the way for more versatile and impactful applications of language agents across a myriad of fields. 


\subsection{Efficiency against Redundant Interactions}
Completing a task necessitates intricate, multi-step interactions with the environment. This process results in extensive and repetitive logs, which have been identified as pivotal for task completion \citep{zhang2023you}. However, due to computational constraints, most studies utilize only a limited number of log steps \citep{park2023generative}. Although recent advancements have expanded the capacity of LLMs to handle extended contexts \citep{xiong2023effective}, conducting inference based on these logs is hampered by the inherently slow speed of autoregressive LLMs. This issue is exacerbated in multi-agent interaction environments, where numerous agents generate a substantial volume of interaction logs. 

To tackle this challenge, one potential solution is to incorporate a memory mechanism for storing and retrieving knowledge from these logs. However, the key challenge lies in exploring effective methods to discern salient knowledge and distill relevant information from the logs. Addressing this challenge is crucial for enhancing the efficiency of inference processes in complex, multi-agent scenarios.

\subsection{Customizable Language Agents}\label{sec:customizable}
LLMs are usually supposed to acquire general language ability and common knowledge through pre-training on large-scale corpora, and then cater to human preferences following instructions through further alignment tuning, including instruction tuning and reinforcement learning from human feedback.
Whereas, users have specialized requirements and individual characteristics. Thus, building a customizable assistant from LLMs is of great importance. 

Existing related studies mostly fall into three general methods:
(i) customizable prompting, often with role or tool specifications. CAMEL \citep{li2023camel} prompted LLM with formatted profiles of human-agent pairs to simulate the workflow of diverse groups of internet users or occupations. MetaAgents \citep{li2023metaagents} prompted the language agent to play a specific role in some certain social context. ExpertPrompting \citep{xu2023expertprompting} proposed to prompt LLMs to solve a problem conditioned on an expert identity profile that is best suited for the problem. RoCo \citep{mandi2023roco} assigned robots with an LLM role to talk on their behalves, generating plans for practical tasks. Customizable ChatGPT has also been announced to comply with specified instructions, extra knowledge, and a combination of skills;\footnote{\url{https://openai.com/blog/introducing-gpts}.}
(ii) customizable training. The gradient updates in the language model can further ensure the customizable alignment. 
Auto-UI \citep{zhang2023you} was trained on the Android UI control domain, achieving stable performance as an autonomous agent. For the communicative agent, Character-LLM \citep{shao2023character} trained the LLMs with profiles and detailed scenes, enabling LLMs to mimic well-known people, like Beethoven;
(iii) customizable model editing. Besides training, editing is an alternative to changing stored knowledge in language agents, which improves the factuality and reliability of a customized assistant. ROME \citep{meng2022locating} and MEMIT \citep{meng2022memit} used the \textit{locate-and-edit} method to correct wrong knowledge. Transformer-Patcher \citep{huang2023transformer} further alleviated the error recurrence by real-time sequential editing. Beyond factual knowledge correctness, PersonalityEdit \citep{mao2023editing} changed the model response to match the Big Five personality traits.

Despite the recent progress, the challenge of developing customizable agents still lies within three folds. Firstly, existing studies mainly focus on methods for practical applications in certain, separate domains. 
However, fewer considerations are oriented to the specific requirements of users.
Secondly, language agent customization requires lightweight, efficient, and low-resource consumption, especially for user-level customization. Different from large-scale, general training, customization peruses effective methods that involve fewer data, partial parameters, or only elaborate designed prompts.
Thirdly, the balance between customization and information security needs to be maintained. 
The user's properties and records (such as age, gender, and medical record) may be exposed to an agent, resulting in a risk of privacy leakage.

\subsection{Scaling up Language Agents}
Multi-agent systems have exhibited social phenomena \citep{park2023generative,wang2023voyager,zhu2023ghost}. Inspired by the observations, recent interest has considered scaling the number of language agents \citep{li2023camel} to form a large-scale language model society. However, computation overhead is still an obstacle when modeling multi-agent communications \citep{xi2023rise}. In the realm of future prospects, the exploration of scaling unveils intriguing possibilities across two pivotal domains. Firstly, there arises a profound curiosity concerning the potential emergence of novel capabilities within a singular agent amidst communication. Secondly, comprehending the implications of scaling, such as personality change and social phenomenon, becomes imperative in empowering language agents to address increasingly complex challenges. Furthermore, this comprehension serves as a linchpin in observing, detecting, and mitigating the risks entailed by potentially harmful behaviors, thereby ensuring the secure and beneficial evolution of these agents for the betterment of society.

\subsection{Safety of Language Agents}\label{sec:safety}
Imagine a near future where intelligent agents are anticipated to seamlessly collaborate with humans and other agents, simplifying daily tasks and interacting with diverse environments. This convenience is accompanied by a significant challenge: ensuring the safety of these agents, especially during prolonged, multi-round interactions. For example, the popular user interface agents designed for web operation \citep{zhou2023webarena} and mobile device control \citep{zhang2023you} may result in privacy leakage and permission abuse.
\citet{shaikh2022second} called for attention to the bias and toxicity in Zero-Shot-CoT reasoning as it tends to significantly induce the model to produce harmful or undesirable output, which may also bring negative effects in language agents.
Effectively addressing the safety challenge demands a multifaceted approach. Firstly, the exploration of more robust and controllable model architectures, coupled with an in-depth understanding of their underlying mechanisms, shows great promise. Delving into the intricacies of agent behavior and enhancing the reliability of their responses are pivotal in this endeavor. Secondly, the rapid evolution of attacks tailored for LLMs necessitates a reevaluation of traditional defense techniques. 

Existing studies concerning LLM safety mainly focus on content safety, such as offensiveness, fairness, and bias of LLM-generated contents \citep{zhang2023safetybench}. As language agents are exposed to multi-turn interactions in distinct environments possibly with operating tools, new safety risks may emerge at a systematic level \citep{xu2023sc,satoevaluating}, ranging from instruction input, environment perception, reasoning process, as well as tool use. We summarize three key properties of agent safety risks, including (i) new attacking types, such as operation attacking by environment injection \citep{liu2023prompt}, tool misuse \citep{fumisusing}, jailbreaking \citep{deng2023jailbreaker,wei2023jailbroken}, and privacy leakage \citep{kim2023propile}; (ii) new attacking surface during the interaction between agent-human, agent-agent, and agent-environment; (iii) complex types of environments, such as operation systems, third-party applications, webpages, and virtual environment.

However, the safety of language languages has been underexplored. The definition of language agent safety has not yet reached an agreement. Novel attack methods, specifically designed for language agents, present unique challenges. Consequently, innovative defense strategies must be developed to mitigate safety risks induced by these sophisticated attacks, particularly in complex environments. This dual focus on building benchmarking resources, enhancing internal safety measures, and fortifying defenses against external threats is paramount for ensuring the secure integration of intelligent agents into our daily lives.

\subsection{Evaluation of Language Agents}
Early studies in NLP mainly focus on assessing a specific ability of models, for example, machine translation, question answering, and summarization \citep{chang2023survey}. The evaluation tends to be dataset-centered, which makes it hard to reflect the model's general ability. In the era of LLMs, more comprehensive benchmark datasets have been released, such as MMLU \citep{hendrycks2020measuring}, BIG-Bench \citep{srivastava2022beyond}, and AGI Eval \citep{zhong2023agieval}. However, the major focus of those benchmark datasets is on the understanding and reasoning abilities of LLMs. Besides, they are mostly single-turn evaluations, which makes it hard to evaluate the planning and decision-making abilities of LLM in distinct environments. 

There is an increasing interest in developing environment-centered evaluation approaches. As language agents are exposed to interactive environments, it remains challenging to evaluate those agents in a volatile environment. Specifically, the measurement of task success might be task-specific and ambiguous. For example, in a system control problem \citep{rawles2023android}, a user instruction can be completed by different trajectories, however, it is hard to annotate all possible ways as gold labels for evaluation. To address the challenge, simulation-based evaluation \citep{wang2023large,yang2023learning,ruan2023identifying} has attracted increasing interest. Execution feedback or external judgment can be used to measure if the task is successful or not. However, execution feedback is not always accessible in every kind of environment, and using external judgment may also include model bias \citep{wang2023large}.

Besides assessing task success rate, it is also critical to consider safety risks as discussed in Section \ref{sec:safety}. Furthermore, as language agents may evolve in the environments, especially in multi-agent communities, how to track and evaluate the agent properties is also a challenge.

\section{Conclusion}
In slightly over a year, CoT techniques have substantially enhanced the reasoning capabilities of LLMs. Going beyond the confines of reasoning tasks in NLP, CoT techniques have been expanded to facilitate the development of language agents. These agents have demonstrated the ability to comprehend language instructions and execute actions in diverse environments. This study meticulously examines the evolution from CoT reasoning to the automation of language agents, offering a comprehensive review and delving into key research topics. These topics include investigating the foundational mechanics underpinning CoT techniques, understanding the paradigm shift associated with CoT, and exploring the emergence of language agents facilitated by CoT techniques. Furthermore, this research delineates several promising avenues for future exploration, including aspects related to generalization, efficiency, customization, scaling, and safety.

\bibliography{custom}
\bibliographystyle{tmlr}




\end{document}